\documentclass[journal]{IEEEtran}
\usepackage{amsmath}
\usepackage{graphicx}
\usepackage{url}
\usepackage{booktabs}
\usepackage{multirow}
\usepackage{amssymb}
\usepackage{cite}
\usepackage[colorlinks=true, linkcolor=blue, citecolor=blue,urlcolor=black]{hyperref}

\ifCLASSINFOpdf
\else
\fi
\begin{document}
%
\title{AWEMixer: Adaptive Wavelet-Enhanced Mixer Network for Long-Term Time Series Forecasting}
%
%
%

\author{Qianyang Li ,
        Xingjun Zhang,
        Peng Tao,
        Shaoxun Wang,
        Yancheng Pan,
        and Jia Wei
\thanks{This work was supported in part by the National Natural Science Foundation of China under Grant 62372366.\textit{(Corresponding author: Xingjun Zhang)}}        
\thanks{Qianyang Li, Xingjun Zhang, Shaoxun Wang and Yancheng Pan was with the School of Computer Science and Technology, Xi'an Jiaotong University, Xi'an 710049, China e-mail: (liqianyang@stu.xjtu.edu.cn, xjzhang@xjtu.edu.cn, shaoxunwang@stu.xjtu.edu.cn and 3120105301@stu.xjtu.edu.cn).}
\thanks{Tao Peng was with the Department of Research and Development, Shandong New Beiyang Information Technology Co., Ltd. WeiHai 264200, China e-mail: ( taopeng@newbeiyng.com).}
\thanks{Jia Wei was with the Department of Computer Science and Technology, Tsinghua University, Beijing 100084, China e-mail: ( weijia4473@mail.tsinghua.edu.cn).}

}

%
%

\markboth{IEEE INTERNET OF THINGS JOURNAL,~Vol.~XX, No.~XX, XX~XX}%
{Shell \MakeLowercase{\textit{et al.}}: Bare Demo of IEEEtran.cls for IEEE Journals}
%



\maketitle

\begin{abstract}
Forecasting long-term time series in IoT environments remains a significant challenge due to the non-stationary and multi-scale characteristics of sensor signals. Furthermore, error accumulation causes a decrease in forecast quality when predicting further into the future. Traditional methods are restricted to operate in time-domain, while the global frequency information achieved by Fourier transform would be regarded as stationary signals leading to blur the temporal patterns of transient events.
We propose AWEMixer, an Adaptive Wavelet-Enhanced Mixer Network including two innovative components: 1) a Frequency Router designs to utilize the global periodicity pattern achieved by Fast Fourier Transform to adaptively weight localized wavelet subband, and 2) a Coherent Gated Fusion Block to achieve selective integration of prominent frequency features with multi-scale temporal representation through cross-attention and gating mechanism, which realizes accurate time-frequency localization while remaining robust to noise.
Seven public benchmarks validate that our model is more effective than recent state-of-the-art models. Specifically, our model consistently achieves performance improvement compared with transformer-based and MLP-based state-of-the-art models in long-sequence time series forecasting. Code is available at https://github.com/hit636/AWEMixer.

\end{abstract}


%
\IEEEpeerreviewmaketitle

\section{Introduction}
%
%
%
%

Due to the extensive application of IoT devices in healthcare \cite{li2024review, adil2024healthcare}, agriculture \cite{kumar2024comprehensive}, industry \cite{DBLP:conf/ijcai/InanL25}, and energy \cite{majhi2024comprehensive}, a large amount of time-series data is generated continuously. The time-series data is nonlinear, high frequency, and multidimensional, which makes it challenging to analyze \cite{siam2025artificial, 10043819,DBLP:journals/iotj/HanLPLC22}. Forecasting models are helpful to support the decision-making, optimal resources, and further predictions. The enhancing of forecasting models is the basis to construct more efficient and reliable IoT system. Therefore, Long-Sequence Time Series Forecasting (LSTF) become a hot research direction to modeling extended and complicated dependencies \cite{kim2025comprehensive, qiu2024tfb, wang2024deep}. However, IoT-generated signals exhibit additional complexities \cite{sadeghi2023internet,DBLP:journals/iotj/ZhangMG24}, including hierarchical multi-scale structures, evolving non-stationarity, and transient high-frequency events such as equipment faults or network anomalies. These localized yet significant patterns are often neglected or oversmoothed by models primarily designed to capture long-term dependencies.

In response, the research community has largely converged on two dominant paradigms. The first, Transformer-based architectures \cite{vaswani2017attention, zhou2021informer, nie2022time}, are theoretically powerful for long-range dependencies but suffer from quadratic complexity and a lack of intrinsic temporal inductive biases. The second, more recent paradigm consists of highly efficient MLP-based and Mixer architectures \cite{zeng2023transformers, DBLP:conf/kdd/EkambaramJNSK23, wang2024timemixer}. These models have achieved state-of-the-art performance with linear complexity, establishing themselves as the new backbone for time series forecasting. However, the limitation of these approaches is that they are mostly applied in the time domain. By considering a time series as merely a sequence of discrete numerical values, these approaches force models to infer complex frequency patterns from scratch. This implicit manner of inducing frequency awareness significantly raises the learning difficulty, which is a severe challenge for the non-stationary signals in IoT scenarios where spectral properties change over time.

This shortcoming motivates the exploration of enhancing time-domain models with explicit introduction of frequency information. As compared with time-domain, frequency-domain analysis offers an informative complementing view by extracting simple oscillatory components from complicated signals. Such extraction process eases the learning process of models, which can capture trends and periodic patterns more straightforwardly.

The possibility of this solution is evidenced by the existing methods which widely involve Fast Fourier Transform (FFT) \cite{duhamel1990fast}. Specifically, FEDformer \cite{zhou2022fedformer} replaces self-attention with an more efficient frequency-domain operator. FiLM \cite{zhou2022film} builds a concise frequency-based representation of the whole historical series. Recently, TimesNet \cite{DBLP:conf/iclr/WuHLZ0L23} uses FFT to find out key periods and reshapes the 1D sequence into 2D format so that a 2D CNN can extract variations within and across periods. All these methods demonstrate that introducing inductive biases in frequency-domain is an effective solution for long-sequence time-series forecasting.

\begin{figure}[t!]
    \centering
    \includegraphics[width=0.85\columnwidth]{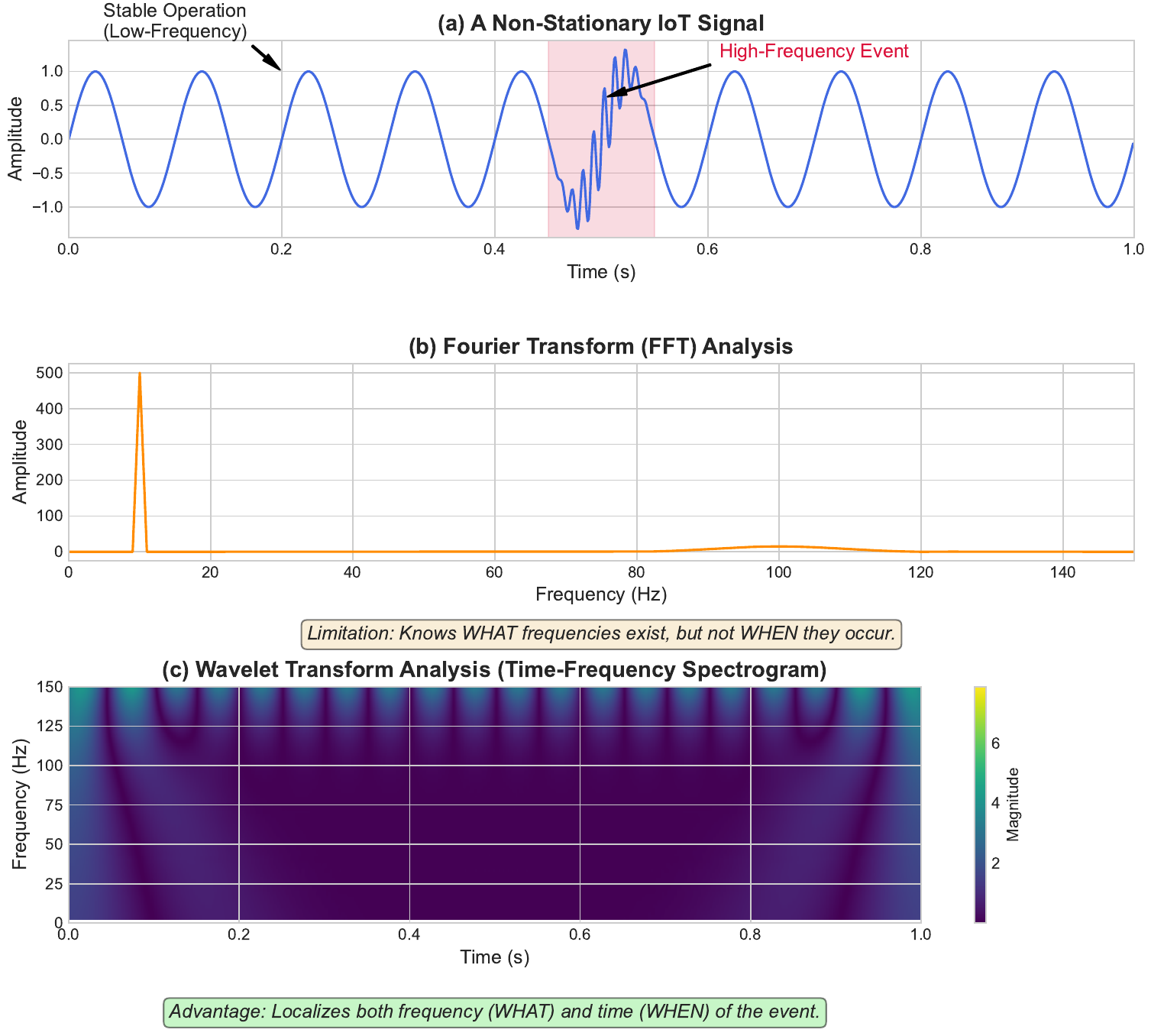}
    \caption{Conceptual illustration. (a) A non-stationary IoT signal with a transient high-frequency event. (b) The Fourier Transform captures the global frequencies but loses the temporal location of the event. (c) The Wavelet Transform localizes the event in both time and frequency, demonstrating its suitability for non-stationary signal analysis.}
    \label{fig:motivation}
\end{figure}

However, the reliance on Fourier Transform is limiting. FFT offers a global view of spectral contents, i.e., it tells which frequencies exist in the time series but provides no information of {when} they appear. However, as shown in Fig. \ref{fig:motivation}, it is the when that holds valuable information in IoT applications since important information is often embedded in transient events. For example, 

\begin{enumerate}

    \item {Transient Faults in Industrial IoT:} A sudden high-frequency oscillation in a sensor reading may indicate impending equipment failure. A global method like FFT would average out the high-frequency event over the entire time window \cite{DBLP:journals/tim/LiLCDSLL23}.
    \item {Dynamic Demand in Smart Grids:} Electricity demand may suddenly change due to an unexpected sudden weather. It thus changes the periodicity of the signal over time \cite{zhou2021informer}.
    \item {Anomalies in Health IoT:} A patient's ECG signal is composed of a periodic heartbeat (which is stable) and potential sudden non-stationary life-threatening arrhythmias (which are transient events) \cite{DBLP:journals/kbs/ZhangTCYX20}.

\end{enumerate}
These applications require models that can localize signal components in both time and frequency. Fourier Transform is inadequate to achieve this. The Wavelet Transform \cite{sundararajan2016discrete}, on the other hand, is a multi-resolution method and is promising to localize non-stationary events.
Recently, methods like WaveletMixer \cite{zhang2025waveletmixer} have demonstrated the benefits of replacing Fourier transforms with Discrete Wavelet Transform (DWT) in Mixer-style architectures. However, since the method is dependent on fixed wavelet bases, it is limited in modeling time series with diverse periodic patterns.

In this paper, we propose AWEMixer that makes two key modifications. First, we replace the predefined wavelets with a set of adaptive wavelet bases whose parameters are directly learned from data. Second, we design a Wavelet Enhancement Module that magnifies the most discriminative frequency components discovered by the adaptive transform. Those components allow the model to construct its own frequency representations for different input time series, leading to better forecasting performance. Our contributions can be outlined in the following points:
\begin{enumerate}
    \item We introduce an adaptive wavelet enhancement mechanism as the backbone of AWEMixer. Two synergistic components are explored: a Frequency Router that dynamically applies wavelet bands over input periodic structure; and a Gated Fusion Block where temporal features attend to the refined frequency representation for selective and periodic robust enhancement.
    \item We adopt a dual-stream architecture that models the signal to obtain two complementary representations from different views: a multi-scale temporal view for trend information and a multi-band wavelet view for local time-frequency patterns. Such a design provides the perfect interaction structure for our adaptive fusion mechanism.
    \item We develop a Cross-Scale Mixing stage that operates on the temporal representation enhanced by wavelet. By allowing information exchange among different temporal scales after they are fused with adaptive frequency features, this mechanism encourages the model to build a holistic representation that comprehends both temporal dynamics and frequency information for the final forecast.
    
    \item Through extensive experiments on multiple public benchmarks, we demonstrate that AWEMixer sets a new state-of-the-art. Our results validate the hypothesis that a forecasting architecture built on adaptive wavelet enhancement is a more powerful and robust paradigm for handling complex, non-stationary time series.
\end{enumerate}

The remainder of this paper is organized as follows. Section \ref{relate} reviews related work. Section \ref{pre} shows the preliminaries. Section \ref{sec:met} details the architecture of AWEMixer. Section \ref{exp} presents our experimental setup and results. Section \ref{sec:discussion} discusses future directions. Finally, Section \ref{con} concludes the paper.
 

\section{Related Work}
\label{relate}

This section reviews the major lines of research in long-term time series forecasting. We organize our discussion around four key architectural paradigms, providing a clear context for the contributions of our proposed model.

\subsection{Transformer-based Temporal Modeling}

Transformers \cite{vaswani2017attention} have become a leading architecture for sequence modeling. Researchers first adapted them for time series in models like the Log-Sparse Transformer \cite{DBLP:conf/nips/LiJXZCWY19}. A major challenge has always been the self-attention mechanism's quadratic complexity, which is a bottleneck for long sequences. Much of the follow-up work aimed to solve this efficiency problem. For example, Informer \cite{zhou2021informer} introduced ProbSparse attention to focus on a smaller set of important queries. Autoformer \cite{wu2021autoformer} took a different route, replacing attention with an auto-correlation mechanism to leverage signal periodicity. More recently, PatchTST \cite{nie2022time} successfully applied the patching technique from Vision Transformers \cite{DBLP:conf/iclr/DosovitskiyB0WZ21} to time series. By dividing the series into patches, it significantly reduces the input sequence length for the Transformer. This work also showed that channel-independent modeling is effective. These models are good at modeling long-range dependencies and they operate primarily in the time domain.

\subsection{Efficient MLP-based Architectures}

Given the high computational overhead, Transformer models have motivated research into simpler models that are more efficient. Several variants have been proposed, and MLP-based models have achieved surprisingly good performance. For example, DLinear model in \cite{zeng2023transformers} adopts an easy linear decomposition into trend and seasonal components but achieves better performance than various Transformer-based methods. Thus, problem formulation and structural inductive bias are more important than model complexity. Subsequent work, e.g., TSMixer \cite{DBLP:conf/kdd/EkambaramJNSK23}, extended the MLP-Mixer \cite{DBLP:conf/nips/TolstikhinHKBZU21} method to allow for the mixing of information across both temporal and feature dimensions, which motivated careful design of the architecture.

\subsection{Frequency-Domain Analysis}

Another perspective is to study time series in the frequency domain. This is particularly suitable when time series exhibit distinct periodic behaviour, and the Fast Fourier Transform (FFT) \cite{duhamel1990fast} is widely used in such scenarios. FEDformer \cite{zhou2022fedformer} adopts attention in the frequency domain and achieves linear complexity, and FiLM \cite{zhou2022film} is almost entirely conducted in the frequency domain. TimesNet \cite{DBLP:conf/iclr/WuHLZ0L23} finds out the major periods via FFT and reshapes the 1D sequence into a 2D tensor so that 2D convolutions can model both intra-period and inter-period variations. However, the main drawback of Fourier-based methods is that they cannot localise transient events in time, which is a serious weakness when applied to non-stationary IoT signals. For example, when there is a system fault or a sudden spike, it is crucial to identify it. Therefore, methods that can model both frequency and temporal information should be adopted. Although the Wavelet Transform has the desired property, its application in deep learning-based forecasting is limited. Even the recent WaveletMixer \cite{zhang2025waveletmixer}, which decomposes signals into multi-resolution components for an ensemble of models, is not perfect. Therefore, AWEMixer proposes an end-to-end network with a dynamic fusion mechanism that can adapt to the input and thus select which parts are more suitable for temporal representations and which parts are more suitable for frequency representations.

\subsection{Multi-Scale and Mixer Paradigms}

Another direction is to model time series explicitly at multiple temporal scales. Such a design is natural given the hierarchical nature of real-world signals. TimeMixer \cite{wang2024timemixer} designs a decomposable multi-scale design, and MSDMixer \cite{MSDMixer} progressively decomposes the series at different layers of the network. MIEL \cite{10948497} proposes a multiscale input ensemble linear network. It utilizes random forward sampling to produce inputs at different scales and thus enhances ultra-long-term forecasting in a single and simple model. All these works have demonstrated that embedding multi-scale reasoning into models is beneficial.

In AWEMixer, we draw inspirations from the above two research streams. We design a lightweight mixing network where a multi-scale temporal hierarchy is introduced. Particularly, we adopt a parallel multi-band frequency representation based on the Wavelet Transform as an additional complementation to the limitation of being purely in the time-domain and global frequency-domain methods. More importantly, AWEMixer enables a dynamic and gated Wavelet Enhancement of temporal features, which makes it a new type of Wavelet-Enhanced Mixing Network for complex and non-stationary time series.

\section{Preliminaries}
\label{pre}
In this section, we first present the problem definition of long-sequence time series forecasting in Section \ref{sec:probdef}. Then, we introduce several basic technical concepts that are the foundation of our proposed AWEMixer architecture in Section \ref{sec:met}.

\subsection{Problem Formulation}
\label{sec:probdef}
The primary objective of long-sequence time series forecasting (LSTF) is to predict a future sequence of values based on a long history of observations. Formally, given a multivariate historical time series of length $L$ with $C$ channels, denoted as $\mathcal{X} = \{\mathbf{x}_1, \dots, \mathbf{x}_L\} \in \mathbb{R}^{L \times C}$, where $\mathbf{x}_t \in \mathbb{R}^{C}$ is the observation at time step $t$, the task is to predict the corresponding future time series of length $T$, denoted as $\mathcal{Y} = \{\mathbf{x}_{L+1}, \dots, \mathbf{x}_{L+T}\} \in \mathbb{R}^{T \times C}$. Our model, following recent state-of-the-art approaches, adopts a channel-independent strategy, treating each of the $C$ channels as a univariate time series to be modeled independently. 

\subsection{Discrete Wavelet Transform (DWT)}
The Discrete Wavelet Transform (DWT) is a signal processing technique that operates by decomposing a signal into a set of basis functions known as wavelets. Unlike the Fourier Transform, which represents signals using sinusoidal functions and loses temporal information, the DWT is capable of simultaneously capturing both frequency content and its timing. This key advantage of time-frequency localization makes it particularly suitable for analyzing non-stationary time series containing transient events such as spikes, shifts, or short-term oscillations.

The DWT operates by recursively passing a signal through pairs of complementary filters: a low-pass filter (LPF) $g$ and a high-pass filter (HPF) $h$. For a discrete signal $x[n]$, the first level of decomposition is mathematically expressed as a convolution followed by a dyadic decimation (downsampling by 2):
\begin{align}
    \mathbf{cA}_1[k] &= \sum_n x[n] \cdot g[2k-n] \\
    \mathbf{cD}_1[k] &= \sum_n x[n] \cdot h[2k-n]
\end{align}
Here, $\mathbf{cA}_1$ are the approximation coefficients, representing the low-frequency, slow-moving trend of the signal. $\mathbf{cD}_1$ are the detail coefficients, capturing the high-frequency, fast-changing details.

For a multi-level decomposition of level $J$, this process is iterated on the approximation coefficients of the previous level. For any level $j \in \{2, \dots, J\}$:
\begin{align}
    \mathbf{cA}_j[k] &= \sum_n \mathbf{cA}_{j-1}[n] \cdot g[2k-n] \\
    \mathbf{cD}_j[k] &= \sum_n \mathbf{cA}_{j-1}[n] \cdot h[2k-n]
\end{align}
This recursive process, known as `wavedec`, results in a hierarchical set of coefficients $[\mathbf{cA}_J, \mathbf{cD}_J, \dots, \mathbf{cD}_1]$, providing a multi-resolution analysis of the signal from its coarsest trend to its finest details.

\subsection{Cross-Attention Mechanism}
The attention mechanism, introduced by Vaswani et al.\cite{vaswani2017attention}, has become a cornerstone of modern deep learning. Its core operation is to compute a weighted sum of values, with the weights configured based on how similar a query is to a collection of keys. The typical definition is :
\begin{equation}
\text{Attention}(\mathbf{Q}, \mathbf{K}, \mathbf{V}) = \text{softmax}\left(\frac{\mathbf{Q}\mathbf{K}^T}{\sqrt{d_k}}\right)\mathbf{V}
\end{equation}
where $\mathbf{Q}$, $\mathbf{K}$, $\mathbf{V}$ represent the matrices of Query, Key, and Value parameters respectively, and $d_k$ represents the dimension of the keys.

\begin{figure*}[htbp]
    \centering
    \includegraphics[width=0.99\textwidth]{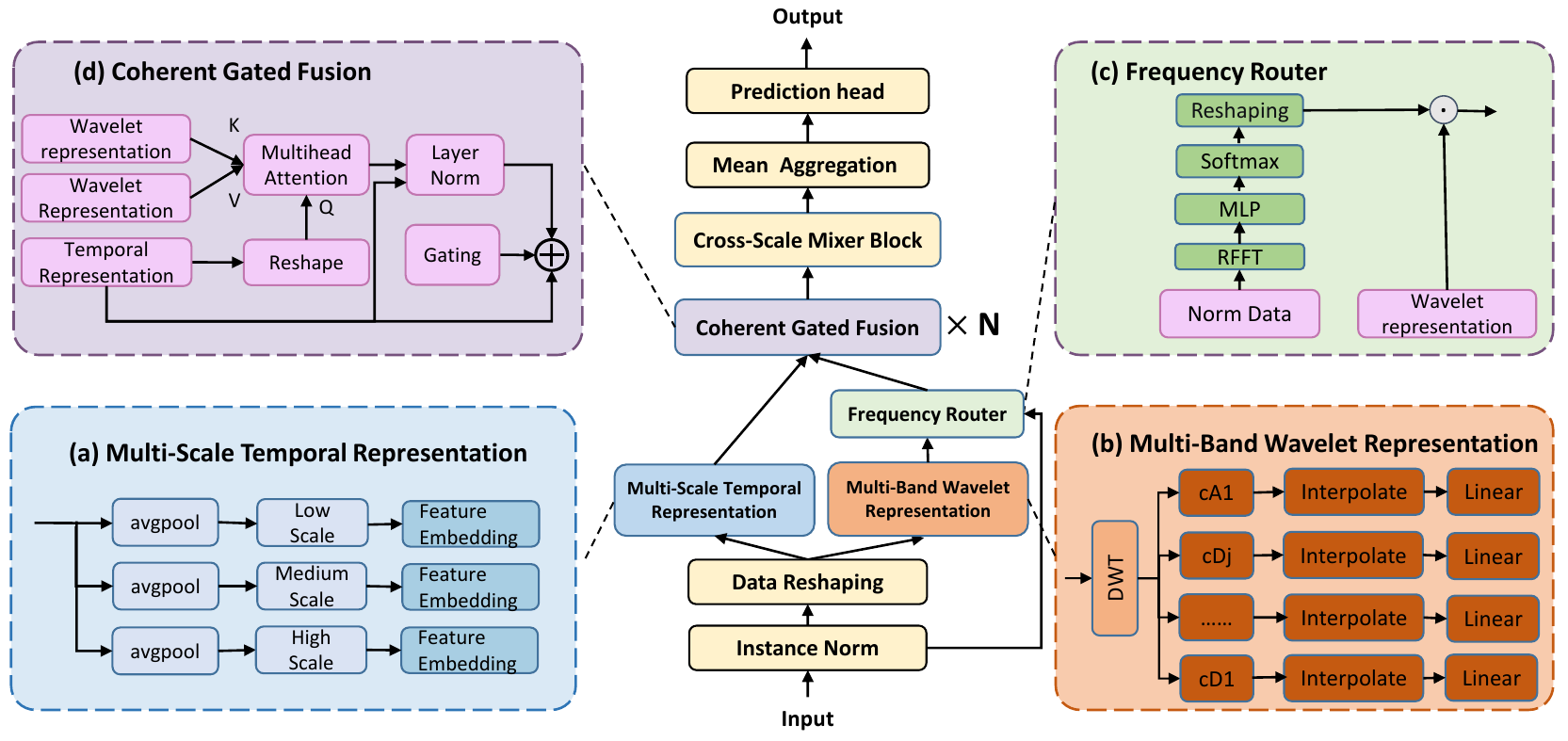} 
    \caption{The overall architecture of AWEMixer. The model consists of a dual-stream backbone for feature extraction. (a) The temporal stream uses hierarchical average pooling to generate a multi-scale representation, capturing features from fine-grained fluctuations to long-term trends. (b) The frequency stream applies a Discrete Wavelet Transform and interpolation to create a parallel multi-band representation. An innovative (c) Frequency Router dynamically weights the wavelet bands based on the input's global periodicity. These adaptive frequency features are then selectively merged with each temporal scale in the (d) Coherent Gated Fusion block, followed by cross-scale mixing before the final prediction.}
    \label{fig:architecture}
\end{figure*}
This work employs cross-attention in which its Query comes from another source different than Keys and Values. Specifically, the temporal features serve as queries to learn and concatenate informative information from frequency-domain signals, thus connecting the two complementary spaces.

\section{Methodology}
\label{sec:met} 
In this section, we present the architecture of AWEMixer. We formulate the model under the assumption that a meaningful time-series analysis should jointly model temporal dynamics and frequency characteristics. As shown in Figure \ref{fig:architecture}, our model is composed of five major pipeline stages hierarchically connected together: input normalization, hierarchical backbone consisting of dual-stream encoder, dynamic frequency routing module, fusion and mixing module, and final prediction head. All the backbone and mixers preserve their linear complexity $O(L)$ with regard to input sequence length $L$, which makes our model scalable to long-sequence forecasting tasks.

\subsection{Input Processing: Reversible Instance Normalization (RevIN)}

First of all, we implement Reversible Instance Normalization \cite{kim2021reversible} to handle distribution shift between time series instances, which usually have very different scales and statistical properties. Specifically, RevinNORM normalizes each instance independently, thereby eliminating the impact of magnitude and location variations while keeping the relative temporal order information.

In case of an input series $\mathbf{x} \in \mathbb{R}^{L}$, RevIN computes its mean $\mu_{\mathbf{x}}$ and standard deviation $\sigma_{\mathbf{x}}$ and subsequently normalizes the series  $\mathbf{x}{\text{norm}} = (\mathbf{x} - \mu{\mathbf{x}}) / (\sigma_{\mathbf{x}} + \epsilon)$, where $\epsilon$ is a small positive constant for numerical stability. The resulting $\mathbf{x}_{\text{norm}} \in \mathbb{R}^{L}$ is passed into the model backbone. A key property of RevIN is its reversibility: after the model produces forecasts, the original statistics restore the predictions to their physically meaningful scale.

\subsection{The Dual-Stream Hierarchical Backbone}

AWEMixer employs a dual-stream backbone to construct two complementary representations: a multi-Scale temporal representation and a multi-Band wavelet-based representation.

\subsubsection{Multi-Scale Temporal Representation}

To capture the hierarchical structure often present in real-world time series, the model explicitly builds a multi-scale temporal representation. Beginning with the normalized input $\mathbf{x}_{\text{norm}}$, a set of $S$ scaled representations is generated through 1D average pooling. For each scale $s \in {0, 1, \dots, S-1}$, pooling is applied with kernel size and stride equal to $2^s$, producing progressively downsampled sequences ${ \mathbf{h}^{(0)}, \mathbf{h}^{(1)}, \dots, \mathbf{h}^{(S-1)} }$. This multi-scale decomposition enables the model to concurrently analyze both short-term fluctuations and long-term trends. Each of these series is then passed through a dedicated, scale-specific linear layer that projects the entire sequence into a single $D$-dimensional feature vector:
\begin{equation}
    \mathbf{z}_t^{(s)} = \text{Linear}_s(\mathbf{h}^{(s)}) \in \mathbb{R}^D
\end{equation}
The output are $S$ sequences of temporal feature vectors $ \{ \mathbf{z}_t^{(0)}, \dots, \mathbf{z}_t^{(S-1)} \}$, where each feature vector captures the information of the time series at a different temporal granularity.

\subsubsection{Multi-Band Wavelet Representation}

Simultaneously, the signal is also transformed in the frequency domain, specifically $J$ levels of Discrete Wavelet Transform are applied to $\mathbf{x}_{\text{norm}}$ resulting in $J+1$ coefficient arrays  $[\mathbf{cA}_J, \mathbf{cD}_J, \dots, \mathbf{cD}_1]$. However due to the dyadic decimation nature of the DWT, the lengths of these arrays are all different. To create a regular representation out of these arrays, we upsample each of them to the original input length $L$ using linear interpolation. We use linear interpolation as an efficient approximation of a continuous function over the discrete wavelet coefficients to obtain an continuous function and subsequently resampling - which is adequate for shifting the bands to a common temporal dimension.  This means all frequency bands are now aligned to a common temporal dimension.
Subsequently, each of the $J+1$ shifted coefficient series is projected into a $D$-dimensional feature space using separate linear embedding layers.
Each of the shifted coefficient series is projected into a D-dimensional feature space using separate linear embedding layers.
\begin{equation}
\mathbf{z}_w^{(j)} = \text{Linear}_j(\text{Interpolate}(\text{coeff}_j)) \in \mathbb{R}^D
\end{equation}
Finally the band embeddings are stacked together to create the initial Multi-Band Wavelet Representation $\mathbf{H}_w\in \mathbb{R}^{N{bands} \times D}$, where $N{bands} = J+1$.  which will be fed through subsequent layers. The tensor $\mathbf{H}_w$\ acts as a structured frequency-domain representation of the signal.

\subsection{Frequency Router: Dynamic Band Weighting}

Global-to-local weighting is implemented by the Frequency Router module which provides information about the relative importance of local patches across different wavelet bands. Although wavelet decomposition provides a localized time-frequency representation of the signal, we first extract a global view of the periodicity patterns present in the sequence using Real Fast Fourier Transform (RFFT). The amplitude spectrum $\mathbf{A} = |\text{RFFT}(\mathbf{x}_{\text{norm}})|$ captures the most prominent frequency components present in the input sequence.

We apply a two-layer MLP with GELU activation on this global frequency profile to learn a set of normalized importance weights over the different wavelet bands. The network first projects the $L/2+1$-dimensional vector representing the amplitude spectrum to a hidden dimension $N_{bands}$ before applying an output projection to dimension $D_{model}$.
\begin{equation}
\mathbf{w} = \text{Softmax}(\text{MLP}(\mathbf{A})) \in \mathbb{R}^{N_{bands}}
\end{equation}
The softmax layer ensures that the weights are a valid probability distribution over the different wavelet bands.

These computed weights $\mathbf{w}$ re-weight the Multi-Band Wavelet Representation $\mathbf{H}_w$ element-wise. Since the weight vector $\mathbf{w}$ is broadcast across the feature dimension, the re-weighting operation can be written as:
\begin{equation}
\mathbf{H}'_w = \mathbf{H}_w \odot \mathbf{w}
\end{equation}
This re-weighting operation yields an adaptively filtered frequency representation $\mathbf{H}'w \in \mathbb{R}^{N{bands} \times D}$, where the resulting bands capture salient global periodic patterns present in the input sequence, while the other less-relevant frequency bands are suppressed. The weighted representation can preserve the original structural organization while re-distributing the information content to the frequency components that best characterize the underlying temporal patterns, enabling more effective cross-domain fusion.

\subsection{Coherent Gated Fusion}

The Coherent Gated Fusion Block constitutes the core strategy applied to enhance the wavelet representation. For each temporal feature vector $\mathbf{z}_t^{(s)}$ and the weighted wavelet representation $\mathbf{H}'_w$, the following operations are applied.

First, a cross-attention layer is applied where the temporal feature vector acts as the query and the wavelet representation acts as keys and values.
\begin{equation}
\text{Con}^{(s)} = \text{Attention}(\mathbf{Q}=\mathbf{z}_t^{(s)}, \mathbf{K}=\mathbf{H}'_w, \mathbf{V}=\mathbf{H}'_w) \in \mathbb{R}^D
\end{equation}

The context vector is concatenated with the raw feature via a residual connection and layer normalization:
\begin{equation}
\mathbf{z}_{\text{enhanced}}^{(s)} = \text{LayerNorm}(\mathbf{z}_t^{(s)} + \text{Con}^{(s)})
\end{equation}

Subsequently, a gate is applied to control the fusion. The gate value is derived from the concatenation of raw and enhanced features:
\begin{equation}
g = \sigma(\text{Linear}([\mathbf{z}t^{(s)}, \mathbf{z}_{\text{enhanced}}^{(s)}]))
\end{equation}

Finally, the output feature is a mixture of enhanced and raw features:
\begin{equation}
\mathbf{z}_{\text{fused}}^{(s)} = g \odot \mathbf{z}_{\text{enhanced}}^{(s)} + (1-g) \odot \mathbf{z}_t^{(s)}
\end{equation}

This procedure is repeated across all temporal scales and can be stacked in multiple layers to refine the fusion depth.

\subsection{Cross-Scale Mixer Block}

After each temporal scale has been independently enriched with frequency information, the Cross-Scale Mixer Block facilitates communication between these different scales. The block takes as input all $S$ fused temporal vectors, stacked into a single tensor $\mathbf{Z}_{\text{fused}} \in \mathbb{R}^{S \times D}$. Inspired by the MLP-Mixer architecture, we apply an MLP across the scale dimension. The tensor is transposed, and an MLP mixes information across the $S$ scales for each of the $D$ feature dimensions. The tensor is then transposed back, with a residual connection for stable training:
\begin{equation}
    \mathbf{Z}_{\text{mixed}} = (\text{MLP}(\mathbf{Z}_{\text{fused}}^T))^T + \mathbf{Z}_{\text{fused}}
\end{equation}
The output is a tensor of mixed-scale representations, $\mathbf{Z}_{\text{mixed}} \in \mathbb{R}^{S \times D}$, where each scale's feature vector now contains contextual information from all other scales.

\subsection{Aggregation and Prediction Head}

The final stage of the model is to aggregate the rich, multi-scale representations and map them to a final forecast. This module takes the mixed-scale tensor $\mathbf{Z}_{\text{mixed}}$ as input. To produce a single, robust summary vector, we apply mean pooling across the scale dimension:
\begin{equation}
    \mathbf{z}_{\text{final}} = \frac{1}{S} \sum_{s=0}^{S-1} \mathbf{Z}_{\text{mixed}}[s,:] \in \mathbb{R}^D
\end{equation}
This final representation vector, $\mathbf{z}_{\text{final}}$, is then passed through a two-layer MLP that serves as the prediction head. This head maps the $D$-dimensional feature vector to the desired forecast horizon, producing the final normalized forecast vector $\hat{\mathbf{y}}_{\text{norm}} \in \mathbb{R}^{T}$.

\section{Experiments}
\label{exp}

In this section, we conduct a systematic evaluation of AWEMixer. First, we compare its forecasting accuracy against current state-of-the-art models on seven public benchmarks. Then, through ablation studies, we analyze the contribution of each key component—particularly the Frequency Router and Gated Fusion mechanism. Finally, we examine the model’s sensitivity to critical hyperparameters to assess its robustness and practical applicability.

\subsection{Experimental Setup}
\subsubsection{Datasets}

We evaluate the proposed method on seven real-world time series datasets that cover diverse domains, sampling frequencies, and temporal characteristics. These datasets collectively represent a wide range of forecasting challenges. A detailed statistical summary is provided in Table \ref{tab:dataset_stats}.

ETT (Electricity Transformer Temperature):
This benchmark consists of four datasets (ETTh1, ETTh2, ETTm1, ETTm2) collected from two electricity transformer stations. Each dataset includes seven operational indicators, such as oil temperature and load, recorded at two sampling intervals—1 hour and 15 minutes. The series exhibit clear daily and seasonal periodicity while also showing strong non-stationarity due to dynamic operational and environmental conditions. These properties make the dataset suitable for evaluating a model’s ability to capture both regular and irregular temporal variations.

Electricity:
This dataset records the hourly electricity consumption (in kW) of 321 clients from 2012 to 2014. Its high dimensionality poses challenges for capturing complex dependencies among multiple correlated variables. The data also display multi-scale seasonal patterns with daily and weekly periodicities, allowing for a comprehensive evaluation of a model’s ability to handle overlapping cyclical structures in forecasting tasks.

Weather:
This dataset contains 21 meteorological indicators (e.g., temperature, humidity, and air pressure) collected every 10 minutes throughout the year 2020. The high sampling frequency and complex nonlinear relationships among variables impose stringent requirements on a model’s ability to capture fine-grained temporal dynamics while maintaining robustness to sensor noise.

Exchange:
This financial dataset records daily exchange rates of eight countries from 1990 to 2016. It is characterized by high volatility, strong non-stationarity, and structural breaks caused by global economic events. These properties make it a challenging testbed for assessing a model’s capability to forecast under low signal-to-noise ratios and weak periodicity conditions.

\begin{table}[h!]
\centering
\caption{STATISTICS OF DATASETS IN THE EXPERIMENT}
\label{tab:dataset_stats}
\begin{tabular}{lcccc}
\toprule
\textbf{Dataset} & \textbf{\#Variates} & \textbf{Timesteps} & \textbf{Frequency} & \textbf{Domain} \\
\midrule
ETTh1 / ETTh2 & 7 & 17,420 & 1-hour & Energy \\
ETTm1 / ETTm2 & 7 & 69,680 & 15-min & Energy \\
Electricity & 321 & 26,304 & 1-hour & Energy \\
Weather & 21 & 52,696 & 10-min & Weather \\
Exchange & 8 & 7,588 & Daily & Economy \\
\bottomrule
\end{tabular}
\end{table}

\subsubsection{Baselines}

For a thorough performance assessment, we compare AWEMixer against nine recent models that cover major architectural paradigms in long-term time series forecasting.

\textit{Transformer-based Models}: This category utilizes self-attention mechanisms to capture long-range dependencies. The following representative models are included:

iTransformer (ICLR 2024): This model inverts the conventional role of time and variate dimensions, treating each individual time series as a single token. This allows the self-attention mechanism to directly model multivariate correlations.

TimeXer (NeurIPS 2024): A powerful Transformer architecture specifically designed to handle exogenous variables. We adapt it for the standard multivariate setting for a fair comparison.

PatchTST (ICLR 2023): A highly effective and efficient model that introduced the concepts of patching (treating sub-series as tokens) and channel independence to the Transformer architecture, achieving SOTA results with reduced complexity.

Crossformer (ICLR 2022): Employs a Two-Stage Attention mechanism to explicitly model both cross-time (temporal) and cross-variate (spatial) dependencies within patches.

FEDformer (ICML 2022): Enhances the Transformer with a frequency-domain attention mechanism. It applies the Fourier Transform to achieve a more efficient, block-wise representation for attention calculation.

\textit{MLP-based Models}: This group represents a recent paradigm shift towards simpler, more efficient architectures that often outperform Transformers by leveraging strong inductive biases like decomposition.

MSD-Mixer (VLDB 2024): A powerful model based on the MLP-Mixer architecture that performs multi-scale decomposition hierarchically within its layers to capture features at different temporal resolutions.

TimeMixer (ICLR 2024): This model employs a decomposable multi-scale mixing architecture, which processes trend and seasonal components through separate pathways.

DLinear (ICML 2022): By combining series decomposition with linear projections, this work established that carefully structured linear models can achieve competitive performance in long-term forecasting.

\textit{CNN-based  Architectures:}

TimesNet (LCLR 2023): This approach identifies principal periods in time series data and restructures the 1D sequence into 2D tensors, enabling 2D convolutions to simultaneously capture patterns within and across periods.

\begin{table*}[t]

\caption{Comprehensive results for the long-term forecasting task. All models were evaluated using a fixed input sequence length of $L=96$ across four prediction horizons ${T \in \{96, 192, 336, 720\}}$. The average of the four prediction lengths is called Avg. The best results are highlighted in \textbf{bold}, while the second-best results are \underline{underlined}.}
\centering
\resizebox{\textwidth}{!}{%
\begin{tabular}{ll rr rr rr rr rr rr rr rr rr rr}
\toprule
\multicolumn{2}{l}{\textbf{Models}} & \multicolumn{2}{c}{\textbf{AWEMixer}} & \multicolumn{2}{c}{\textbf{MSD-Mixer}}  &\multicolumn{2}{c} {\textbf{TimeMixer}} & \multicolumn{2}{c}{\textbf{TimeXer}} & \multicolumn{2}{c}{\textbf{iTransformer}} & \multicolumn{2}{c}{\textbf{PatchTST}} & \multicolumn{2}{c}{\textbf{Crossformer}} & \multicolumn{2}{c}{\textbf{TimesNet}} & \multicolumn{2}{c}{\textbf{DLinear}}  & \multicolumn{2}{c}{\textbf{FEDformer}} \\
\multicolumn{2}{l}{} & \multicolumn{2}{c}{\textbf{(Ours)}} & \multicolumn{2}{c}{\textbf{2024}} & \multicolumn{2}{c}{\textbf{2024}}  & \multicolumn{2}{c}{\textbf{2024}} & \multicolumn{2}{c}{\textbf{2024}}& \multicolumn{2}{c}{\textbf{2023}}  & \multicolumn{2}{c}{\textbf{2023}} & \multicolumn{2}{c}{\textbf{2023}} & \multicolumn{2}{c}{\textbf{2023}} & \multicolumn{2}{c}{\textbf{2022}}  \\
\cmidrule(lr){3-4} \cmidrule(lr){5-6} \cmidrule(lr){7-8} \cmidrule(lr){9-10} \cmidrule(lr){11-12} \cmidrule(lr){13-14} \cmidrule(lr){15-16} \cmidrule(lr){17-18} \cmidrule(lr){19-20} \cmidrule(lr){21-22} 
\textbf{Metric} & & \textbf{MSE} & \textbf{MAE} & \textbf{MSE} & \textbf{MAE} & \textbf{MSE} & \textbf{MAE} & \textbf{MSE} & \textbf{MAE} & \textbf{MSE} & \textbf{MAE} & \textbf{MSE} & \textbf{MAE} & \textbf{MSE} & \textbf{MAE} & \textbf{MSE} & \textbf{MAE} & \textbf{MSE} & \textbf{MAE} & \textbf{MSE} & \textbf{MAE}   \\
\midrule
\multirow{5}{*}{Weather} 
& 96 & 0.164 & \textbf{0.203} & \textbf{0.148} & 0.212 & 0.202 & 0.261 & \underline{0.157} & \underline{0.205} & 0.174 & 0.212 & 0.186 & 0.227 & 0.195 & 0.271 & 0.172 & 0.220 & 0.195 & 0.252 & 0.217 & 0.296 \\
& 192 & \textbf{0.199} & \textbf{0.246} & \underline{0.200} & 0.262 & 0.208 & 0.250 & {0.204} & \underline{0.247} & 0.221 & 0.254 & 0.234 & 0.265 & 0.209 & 0.277 & 0.219 & 0.261 & 0.237 & 0.295 & 0.276 & 0.336 \\
& 336 & \textbf{0.250} & {0.292} & 0.256 & 0.310 & \underline{0.251} & \textbf{0.278} & 0.261 & \underline{0.29} & 0.278 & 0.296 & 0.284 & 0.301 & 0.273 & 0.332 & 0.280 & 0.306 & 0.282 & 0.331 & 0.339 & 0.380 \\
& 720 & \underline{0.331} & \textbf{0.339} & \textbf{0.327} & 0.362 & 0.339 & \underline{0.341} & 0.340 & \underline{0.341} & 0.358 & 0.347 & 0.356 & 0.349 & 0.379 & 0.401 & 0.365 & 0.359 & 0.359 & 0.345 & 0.403 & 0.428 \\
\cmidrule(l){2-22}
& Avg & \underline{0.238} & \textbf{0.270} & \textbf{0.233} & 0.286 & 0.250 & 0.283 & \underline{0.241} & {0.271} & 0.258 & 0.278 & 0.265 & 0.285 & 0.264 & 0.320 & 0.259 & 0.287 & 0.265 & 0.315 & 0.309 & 0.360 \\
\midrule
\multirow{5}{*}{Exchange} 
& 96 & \textbf{0.084} & \textbf{0.201} & \underline{0.085} & \underline{0.203} & 0.090 & 0.235 & 0.085 & 0.204 & 0.086 & 0.206 & 0.088 & 0.205 & 0.256 & 0.367 & 0.107 & 0.234 & 0.088 & 0.218 & 0.148 & 0.278 \\
& 192 & \textbf{0.175} & \textbf{0.293} & \underline{0.176} & \underline{0.297} & 0.187 & 0.343 & 0.181 & 0.302 & 0.177 & 0.299 & 0.176 & 0.299 & 0.470 & 0.509 & 0.226 & 0.344 & 0.176 & 0.315 & 0.271 & 0.315 \\
& 336 & \underline{0.335} & {0.431} & {0.336} & {0.418} & 0.353 & 0.473 & 0.363 & 0.435 & \textbf{0.331} & \underline{0.417} & 0.301 & \textbf{0.397} & 1.268 & 0.883 & 0.367 & 0.448 & 0.313 & 0.427 & 0.460 & 0.427 \\
& 720 & \underline{0.894} & \textbf{0.685} & 0.953 & 0.738 & 0.934 & 0.761 & 0.930 & 0.727 & \textbf{0.847} & \underline{0.691} & 0.901 & 0.714 & 1.767 & 1.088 & 0.964 & 0.746 & 0.839 & 0.695 & 1.195 & 0.695 \\
\cmidrule(l){2-22}
& Avg & 0.371 & \textbf{0.402} & 0.388 & 0.414 & 0.391 & 0.453 & 0.389 & 0.417 & \textbf{0.360} & \underline{0.403} & \underline{0.367} & 0.404 & 0.940 & 0.707 & 0.416 & 0.443 & 0.354 & 0.414 & 0.519 & 0.429 \\
\midrule
\multirow{5}{*}{Electricity} 
& 96 & \underline{0.152} & \underline{0.245} & 0.161 & 0.253 & {0.153} & 0.247 & 0.182 & 0.278 & \textbf{0.148} & \textbf{0.240} & 0.190 & 0.296 & 0.219 & 0.314 & 0.168 & 0.272 & 0.210 & 0.305 & 0.169 & 0.273 \\
& 192 & \textbf{0.161} & \textbf{0.254} & 0.169 & 0.269 & \underline{0.166} & {0.256} & 0.199 & 0.263 & \underline{0.162} & 0.253 & 0.196 & 0.304 & 0.231 & 0.322 & 0.184 & 0.298 & 0.210 & 0.305 & 0.201 & 0.315 \\
& 336 & \textbf{0.176} & \textbf{0.275} & 0.188 & 0.286 & \underline{0.185} & 0.277 & 0.193 & 0.312 & \underline{0.178} & \underline{0.269} & 0.217 & 0.319 & 0.246 & 0.337 & 0.198 & 0.300 & 0.223 & 0.319 & 0.200 & 0.304 \\
& 720 & \textbf{0.222} & \textbf{0.308} & 0.229 & \underline{0.311} & 0.225 & 0.317 & 0.233 & 0.312 & \underline{0.225} & \underline{0.311} & 0.258 & 0.352 & 0.280 & 0.363 & \underline{0.220} & \underline{0.320} & 0.258 & 0.350 & 0.246 & 0.355 \\
\cmidrule(l){2-22}
& Avg & \textbf{0.178} & \textbf{0.270} & 0.187 & 0.280 & \underline{0.182} & 0.274 & 0.202 & 0.290 & \underline{0.178} & \underline{0.271} & 0.216 & 0.318 & 0.244 & 0.334 & 0.192 & 0.304 & 0.225 & 0.319 & 0.214 & 0.327 \\
\midrule
\multirow{5}{*}{ETTh1} 
& 96 & \textbf{0.374} & \textbf{0.391} & 0.377 & \underline{0.392} & \underline{0.375} & 0.400 & 0.382 & 0.403 & 0.386 & 0.405 & 0.460 & 0.447 & 0.423 & 0.448 & 0.384 & 0.402 & 0.407 & 0.412 & 0.395 & 0.424 \\
& 192 & \textbf{0.421} & \textbf{0.425} & \underline{0.427} & 0.422 & 0.429 & 0.421 & 0.429 & 0.453 & 0.441 & 0.512 & 0.477 & 0.429 & 0.471 & 0.474 & 0.436 & 0.446 & 0.441 & 0.411 & 0.469 & 0.470 \\
& 336 & \underline{0.466} & \textbf{0.442} & 0.469 & \underline{0.443} & \textbf{0.458} & 0.448 & 0.468 & 0.448 & 0.487 & 0.458 & 0.546 & 0.496 & 0.496 & 0.470 & 0.491 & 0.491 & 0.469 & 0.489 & 0.547 & 0.495 \\
& 720 & \textbf{0.467} & \underline{0.469} & 0.485 & 0.475 & 0.498 & 0.482 & \underline{0.469} & \textbf{0.461} & 0.503 & 0.491 & 0.544 & 0.517 & 0.653 & 0.621 & 0.521 & 0.500 & 0.513 & 0.510 & 0.598 & 0.544 \\
\cmidrule(l){2-22}
& Avg & \textbf{0.432} & \textbf{0.431} & \underline{0.440} & \underline{0.433} & 0.440 & 0.438 & \underline{0.437} & 0.437 & 0.454 & 0.447 & 0.516 & 0.484 & 0.529 & 0.522 & 0.458 & 0.450 & 0.461 & 0.457 & 0.498 & 0.484 \\
\midrule
\multirow{5}{*}{ETTh2} 
& 96 & \textbf{0.282} & \textbf{0.336} & \underline{0.284} & 0.345 & 0.289 & 0.342 & 0.308 & 0.355 & {0.286} & \underline{0.338} & 0.745 & 0.584 & 0.400 & 0.440 & 0.340 & 0.374 & 0.340 & 0.394 & 0.358 & 0.397 \\
& 192 & \textbf{0.361} & \textbf{0.387} & \underline{0.362} & 0.392 & 0.378 & \underline{0.397} & 0.363 & 0.389 & 0.380 & 0.400 & 0.793 & 0.585 & 0.877 & 0.656 & 0.402 & 0.452 & 0.419 & 0.479 & 0.414 & 0.439 \\
& 336 & \underline{0.395} & \underline{0.421} & 0.399 & 0.428 & 0.428 & \textbf{0.386} & \textbf{0.414} & 0.414 & 0.423 & 0.428 & 0.927 & 0.643 & 1.043 & 0.731 & 0.452 & 0.482 & 0.591 & 0.541 & 0.496 & 0.487 \\
& 720 & \textbf{0.411} & \textbf{0.431} & \underline{0.426} & 0.457 & 0.412 & 0.434 & \underline{0.414} & \underline{0.432} & 0.427 & 0.445 & 1.043 & 0.636 & 1.104 & 0.763 & 0.462 & 0.468 & 0.661 & 0.661 & 0.463 & 0.474 \\
\cmidrule(l){2-22}
& Avg & \textbf{0.363} & \textbf{0.393} & 0.368 & 0.406 & \underline{0.364} &{0.398} & 0.383 & \underline{0.396} & 0.383 & 0.407 & 0.878 & 0.612 & 0.841 & 0.642 & 0.414 & 0.427 & 0.563 & 0.519 & 0.437 & 0.449 \\
\midrule
\multirow{5}{*}{ETTm1} 
& 96 & \underline{0.316} & \underline{0.354} & \textbf{0.304} & \textbf{0.351} & 0.320 & 0.357 & 0.318 & 0.356 & 0.334 & 0.368 & 0.352 & 0.374 & 0.404 & 0.426 & 0.338 & 0.375 & 0.346 & 0.374 & 0.379 & 0.419 \\
& 192 & \underline{0.359} & \underline{0.381} & \textbf{0.344} & \textbf{0.375} & 0.361 & 0.383 & 0.362 & 0.383 & 0.404 & 0.393 & 0.387 & 0.404 & 0.450 & 0.451 & 0.374 & 0.387 & 0.381 & 0.391 & 0.389 & 0.387 \\
& 336 & \underline{0.391} & 0.402 & \textbf{0.370} & \textbf{0.395} & 0.390 & 0.404 & 0.395 & 0.407 & 0.426 & 0.420 & 0.421 & 0.414 & 0.532 & 0.515 & 0.410 & 0.411 & 0.415 & 0.415 & 0.445 & 0.459 \\
& 720 & \underline{0.443} & \underline{0.438} & \textbf{0.427} & \textbf{0.428} & 0.454 & 0.441 & 0.441 & 0.491 & 0.491 & 0.459 & 0.462 & 0.449 & 0.666 & 0.589 & 0.478 & 0.450 & 0.473 & 0.451 & 0.543 & 0.490 \\
\cmidrule(l){2-22}
& Avg & \underline{0.377} & \underline{0.393} & \textbf{0.361} & \textbf{0.387} & 0.381 & 0.395 & 0.382 & 0.391 & 0.407 & 0.410 & 0.406 & 0.407 & 0.513 & 0.495 & 0.400 & 0.404 & 0.404 & 0.408 & 0.448 & 0.452 \\
\midrule
\multirow{5}{*}{ETTm2} 
& 96 & \textbf{0.168} & \underline{0.257} & \underline{0.169} & {0.259} & 0.175 & 0.258 & 0.171 & \textbf{0.256} & 0.180 & 0.264 & 0.183 & 0.270 & 0.287 & 0.366 & 0.187 & 0.267 & 0.193 & 0.286 & 0.203 & 0.287 \\
& 192 & \underline{0.236} & \textbf{0.296} & \textbf{0.232} & \underline{0.300} & 0.237 & 0.299 & 0.237 & 0.299 & 0.250 & 0.309 & 0.255 & 0.314 & 0.414 & 0.492 & 0.249 & 0.309 & 0.284 & 0.361 & 0.325 & 0.366 \\
& 336 & \textbf{0.291} & \textbf{0.336} & \underline{0.292} & \underline{0.337} & 0.298 & 0.340 & 0.296 & \underline{0.338} & 0.311 & 0.348 & 0.309 & 0.347 & 0.597 & 0.542 & 0.321 & 0.331 & 0.382 & 0.429 & 0.325 & 0.366 \\
& 720 & \textbf{0.397} & \textbf{0.391} & \underline{0.398} & 0.412 & 0.391 & 0.392 & 0.392 & \underline{0.394} & 0.407 & 0.412 & 0.404 & 0.407 & 1.730 & 1.042 & 0.408 & 0.403 & 0.558 & 0.525 & 0.421 & 0.415 \\
\cmidrule(l){2-22}
& Avg & \textbf{0.273} & \textbf{0.320} & \underline{0.272} & \underline{0.324} & 0.275 & 0.323 & 0.274 & \underline{0.322} & 0.288 & 0.332 & 0.290 & 0.334 & 0.757 & 0.610 & 0.291 & 0.333 & 0.354 & 0.402 & 0.305 & 0.349 \\
\midrule
\multicolumn{2}{l|}{1st Count} & \textbf{20} & \textbf{24} & 9 & 5 & 1 & 2 & 1 & 2 & 4 & 1 & 0 & 1 & 0 & 0 & 0 & 0 & 0 & 0 & 0 & 0 \\
\bottomrule
\end{tabular}
}

\label{tab:main_results}
\end{table*}

\subsubsection{Implementation Details}

All the experiments are conducted according to the long-term forecasting process in TimesNet \cite{DBLP:conf/iclr/WuHLZ0L23}. We set the input sequence length L=96 for all models and datasets. For forecasting, we need to predict at four horizons $T \in {96, 192, 336, 720}$ to capture the forecastability of short-, medium-, and long-term. The baseline results are extracted from the original papers or public benchmarks with the exact experimental setup following the same comparison principle with the best performance reported in the paper.

AWEMixer is implemented in PyTorch, with PyWavelets \cite{DBLP:journals/jossw/LeeGWWO19} employed for the discrete wavelet transform. The wavelet stream uses a 3-level decomposition with Daubechies 4 ('db4') wavelets. The architecture stacks 3 encoder layers, with model dimension $d_{\text{model}}=128$ in each Gated Fusion Block. The cross-attention mechanism uses 8 attention heads.. We use the Adam optimizer with an initial learning rate of $1 \times 10^{-4}$ and a batch size of 32. All experiments were conducted on a single NVIDIA RTX 4090 GPU. Following state-of-the-art practices, we employ RevIN for reversible instance normalization in all experiments.

\subsubsection{Evaluation Metrics}
We use two standard and complementary metrics to evaluate forecasting performance: Mean Squared Error (MSE) and Mean Absolute Error (MAE). MSE, by squaring the errors, penalizes larger deviations more heavily and is sensitive to outliers. MAE provides a more direct and easily interpretable measure of the average error magnitude. For a prediction $\hat{Y}$ and ground truth $Y$ over a horizon of length $H$, they are defined as:
\begin{align}
     \text{MSE} = \frac{1}{H} \sum_{i=1}^{H} (\hat{Y}_i - Y_i)^2 \\
     \text{MAE} = \frac{1}{H} \sum_{i=1}^{H} |\hat{Y}_i - Y_i|
\end{align}
   
For both metrics, lower values indicate better forecasting accuracy.

\subsection{Main Results}

The comprehensive quantitative results of our long-term forecasting experiments are presented in Table \ref{tab:main_results}. The table details the MSE and MAE for all eleven models across the seven datasets and four prediction horizons, along with the average performance over the horizons.

The results unequivocally establish that AWEMixer achieves a new state-of-the-art performance, demonstrating consistent superiority across the majority of benchmark settings. To evaluate model performance systematically, we employ a "1st Count" metric that tracks how often each model achieves the top result for each evaluation metric across all 35 experimental settings (7 datasets × 5 horizons). Among 70 possible top rankings, AWEMixer attains 20 for MSE and 24 for MAE. This result shows a clear improvement over other individual models, and confirms the advantages of the wavelet mixing approach.

On average over the four horizons ("Avg" rows), this confirms the good performance of AWEMixer. Indeed, it is always the first or second performer in both MSE and MAE on all seven datasets, which shows good generalization ability over different datasets types and prediction horizons.. For instance, on the challenging ETTh2 dataset, known for its non-stationarity, AWEMixer achieves the best average MSE (0.363) and MAE (0.393). On Electricity, a dataset characterized by high dimensionality and strong periodicities, it also delivers the best average performance in both MSE (0.178) and MAE (0.270).

The advantage of AWEMixer is particularly pronounced when contrasted with models from different architectural paradigms. Compared to the suite of powerful Transformer-based models, AWEMixer consistently demonstrates a significant performance advantage. For example, on the ETTh1 average results, AWEMixer's MSE of 0.432 is substantially better than PatchTST's (0.516) and iTransformer's (0.454). This provides strong evidence that for these time series benchmarks, the specific inductive bias provided by our wavelet-enhanced time-frequency fusion is more effective than the general sequence modeling capability of self-attention.

The most crucial comparison is against other decomposition-based models, which represent the current state-of-the-art. AWEMixer consistently outperforms DLinear and TimeMixer. Its main competitor is the powerful MSD-Mixer. While MSD-Mixer shows exceptional performance, especially on the Weather dataset, AWEMixer's dominant performance across the entire ETT family and the volatile Exchange dataset gives it a decisive overall edge. Note the clear difference in design: MSD-Mixer decomposes its input implicitly in the time domain, whereas AWEMixer's explicit decomposition in the time-frequency domain offers a more expressive, flexible, and thus more effective representation for forecasting.

\subsection{Visualizing and Analyzing}
\begin{figure*}[t]
    \centering 

    \begin{minipage}[b]{0.48\columnwidth}
        \centering
        \includegraphics[width=\linewidth]{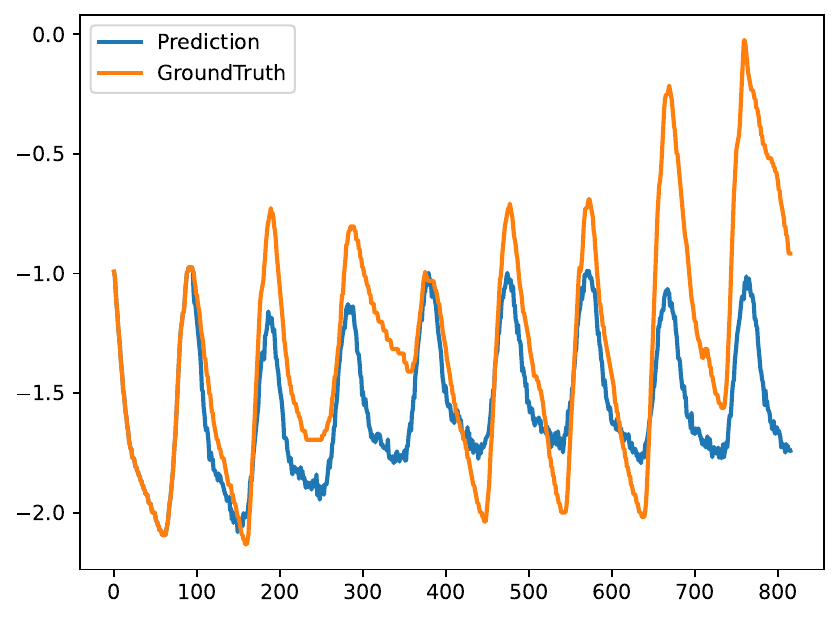}
        \centerline{(a) ETTm2-720-AWEMixer}
    \end{minipage}%
    \hfill
    \begin{minipage}[b]{0.48\columnwidth}
        \centering
        \includegraphics[width=\linewidth]{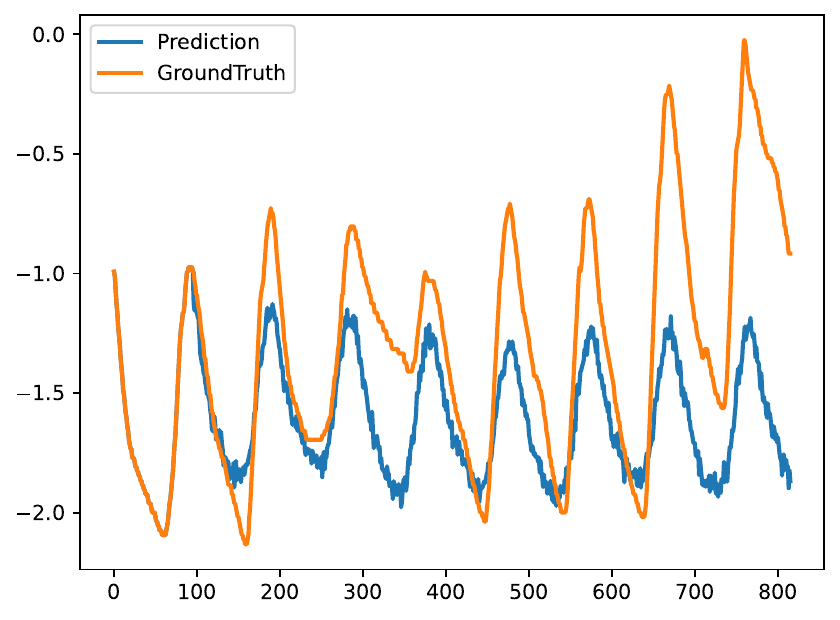}
        \centerline{(b) ETTm2-720-MSDMixer}
    \end{minipage}
     \begin{minipage}[b]{0.48\columnwidth}
            \centering
            \includegraphics[width=\linewidth]{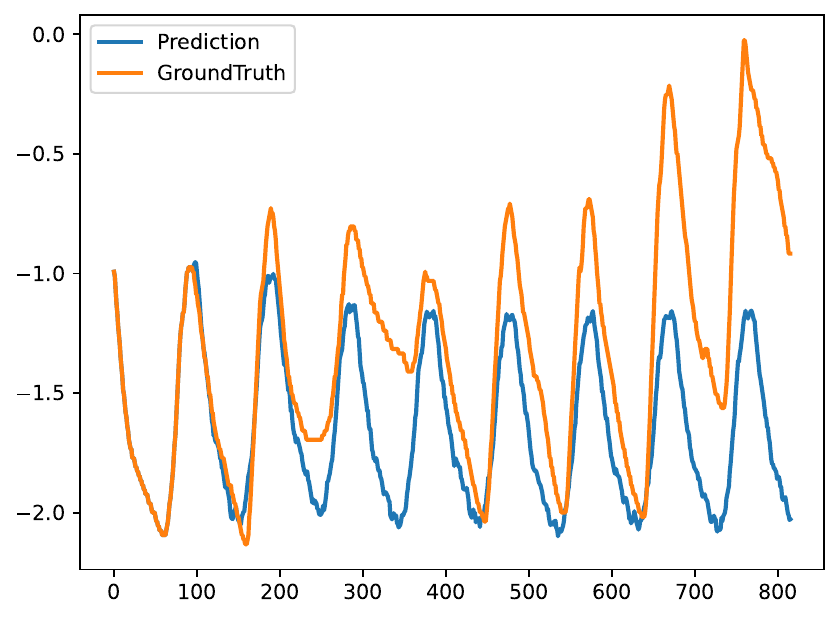}
            \centerline{(c) ETTm2-720-TimeMixer}
        \end{minipage}%
        \hfill
        \begin{minipage}[b]{0.48\columnwidth}
            \centering
            \includegraphics[width=\linewidth]{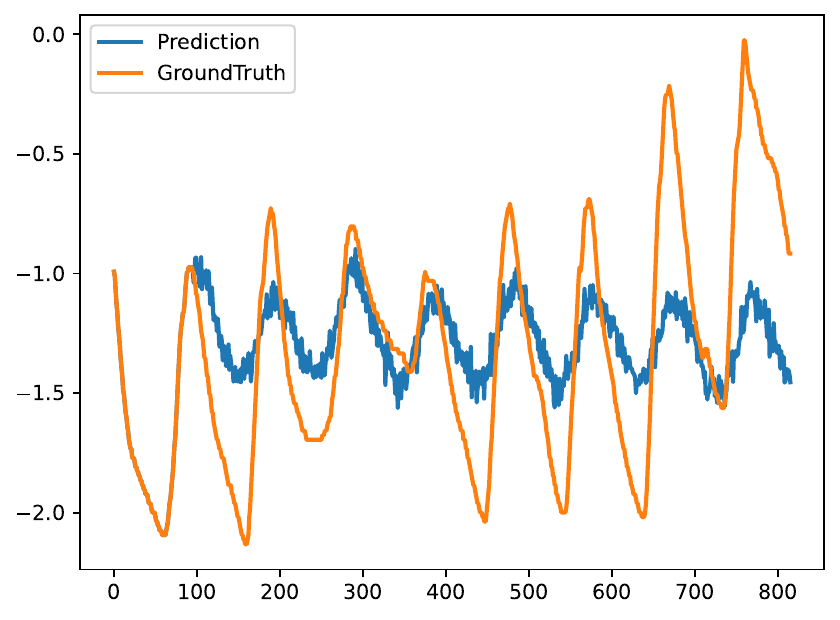}
            \centerline{(d) ETTm2-720-iTransformer}
        \end{minipage}
\vspace{0.3cm} 

    \begin{minipage}[b]{0.48\columnwidth}
        \centering
        \includegraphics[width=\linewidth]{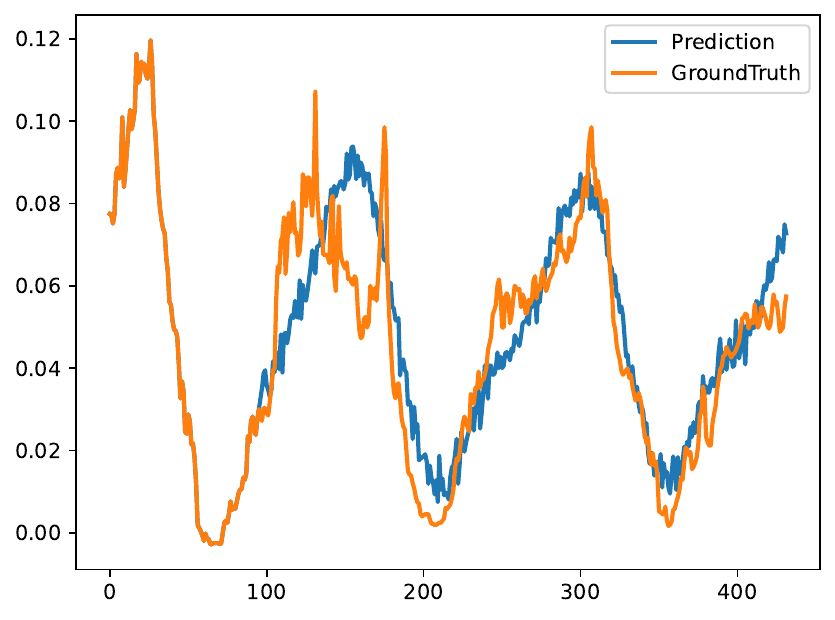}
        \centerline{(e) Weather-336-AWEMixer}
    \end{minipage}%
    \hfill
    \begin{minipage}[b]{0.48\columnwidth}
        \centering
        \includegraphics[width=\linewidth]{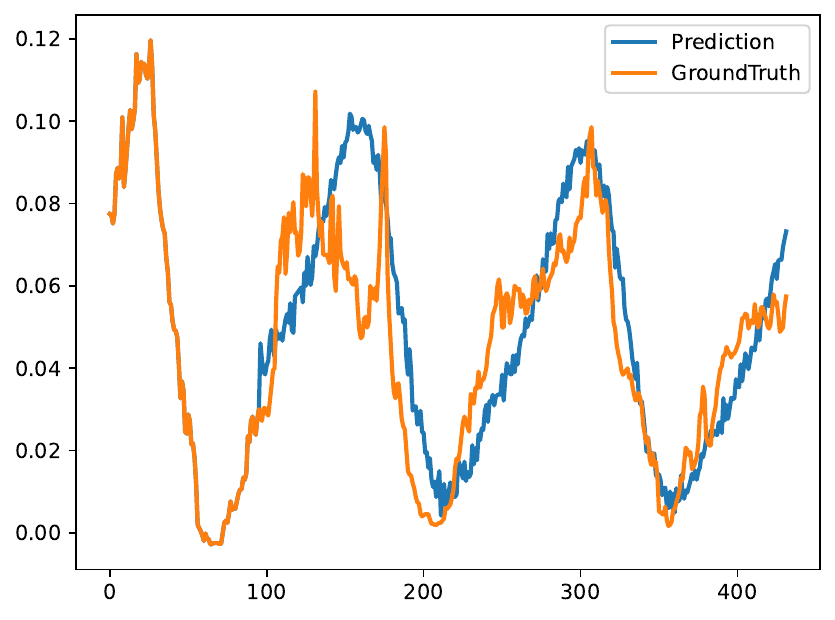}
        \centerline{(f) Weather-336-MSDMixer}
    \end{minipage}
     \begin{minipage}[b]{0.48\columnwidth}
            \centering
            \includegraphics[width=\linewidth]{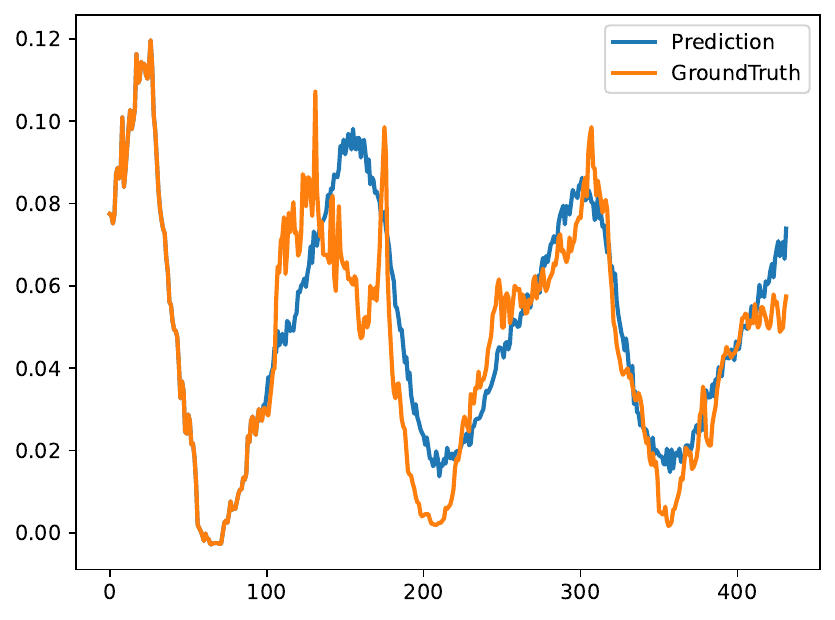}
            \centerline{(g) Weather-336-TimeMixer}
        \end{minipage}%
        \hfill
        \begin{minipage}[b]{0.48\columnwidth}
            \centering
            \includegraphics[width=\linewidth]{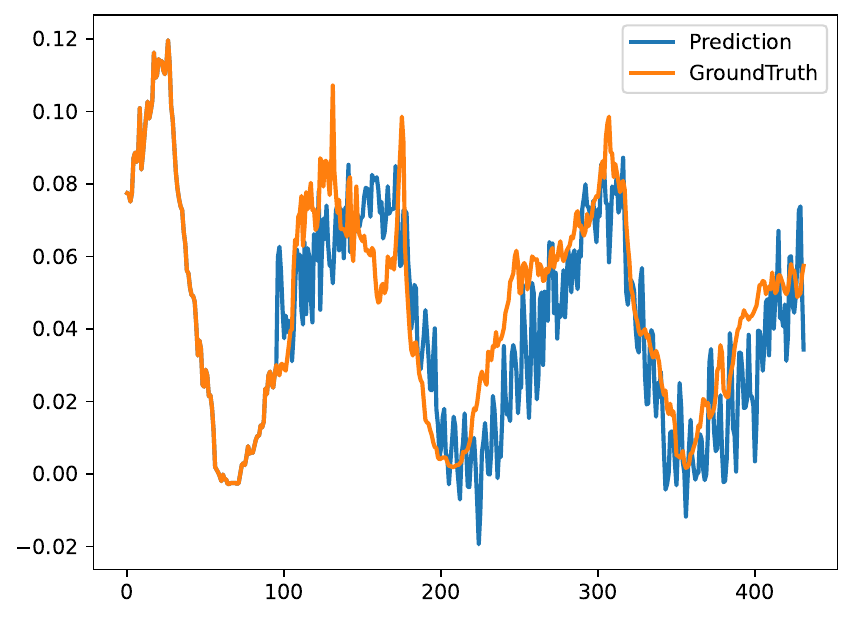}
            \centerline{(h) Weather-336-iTransformer}
        \end{minipage}

\caption{
    Case study comparing the forecasting performance of AWEMixer against baseline models on the ETTm2 and Weather datasets. 
    Each plot displays the ground truth (orange) and a model's prediction (blue). 
    The top row ((a)-(d)) showcases performance on the periodic ETTm2 dataset with a 720-step forecast horizon. 
    The bottom row ((e)-(h)) illustrates robustness on the volatile Weather dataset with a 336-step forecast horizon.
}
    \label{fig:visual}
\end{figure*}

In addition to quantitative evaluation, we visualize AWEMixer's forecasting behavior on long-range forecasting to provide qualitative insights. We evaluate long-term settings in previous works: {ETTm2} with 720 points (7.5 days) forecast and Weather with 336 points (14 days) forecast. Representative examples from two datasets are shown in Figure \ref{fig:visual} compared with ground truth..

\textit{Performance on Periodic Series (ETTm2):}  In Figure \ref{fig:visual} (a–d), we show the forecasting results on ETTm2 with strong periodicity. In (a), AWEMixer (blue) and ground truth (orange) are very close in both period and amplitude. MSDMixer (b) and TimeMixer (c) predict sequences with smaller amplitude and fail to capture the peak value. iTransformer (d) is shifted in phase and the waveform is too smooth, which makes it unable to capture the fine-grained temporal structure. This further demonstrates the ability of AWEMixer to model periodic signals.

\textit{Robustness to Volatility and Noise (Weather):} In the bottom row of Figure \ref{fig:visual} (e–h), we visualize a volatile and noisy part of Weather. AWEMixer (e) predicts a stable sequence, and its output contains the main trend and turning points of the ground truth while removing the high-frequency component. iTransformer (h) is very noisy and the prediction is very irregular, which shows that it is sensitive to short-term fluctuations. Although (f) and (g) maintained the overall trend of Weather, AWEMixer (e) provides a cleaner trend compared with the ground truth, which is due to the noise suppression ability brought by the wavelet effect of AWEMixer.

Collectively, these visualizations qualitatively support the quantitative findings, , i.e., AWEMixer provides temporally coherent and robust representations for periodic signals and noisy sequences.

\subsection{Ablation Studies}

We conduct ablative experiments to study the effectiveness of different components in AWEMixer. We construct four models by sequentially removing one important component in AWEMixer. The performance of these models is evaluated on Weather and ETTm2 by average MSE decrease on all horizon (Table \ref{tab:ablation}).

\begin{table*}[t]
\caption{Ablation study of AWEMixer components on the Weather and ETTh2 datasets. We report MSE and MAE metrics and the average performance degradation (\%) for each ablated component.}
\label{tab:ablation}
\centering
\resizebox{1.0\textwidth}{!}{
\setlength{\tabcolsep}{0.9em}
\begin{tabular}{llcccccccccc}
\toprule
\multicolumn{2}{c}{\textbf{Models}} & \multicolumn{2}{c}{\textbf{Ours}} & \multicolumn{2}{c}{\textbf{w/o Router}} & \multicolumn{2}{c}{\textbf{w/o Wavelet}} & \multicolumn{2}{c}{\textbf{w/o Gating}} & \multicolumn{2}{c}{\textbf{w/o Mixer}}  \\
\cmidrule(lr){3-4} \cmidrule(lr){5-6} \cmidrule(lr){7-8} \cmidrule(lr){9-10} \cmidrule(lr){11-12} 
\textbf{Metric} & & MSE & MAE & MSE & MAE & MSE & MAE & MSE & MAE & MSE & MAE \\
\midrule
\multirow{4}{*}{Weather} 
& 96  & \textbf{0.164} & \textbf{0.203} & 0.171 & 0.209 & 0.169 & 0.208 & 0.168 & 0.212 & 0.179 & 0.231  \\
& 192 & \textbf{0.199} & \textbf{0.246} & 0.208 & 0.257 & 0.211 & 0.259 & 0.209 & 0.265 & 0.223 & 0.276  \\
& 336 & \textbf{0.250} & \textbf{0.292} & 0.263 & 0.305 & 0.269 & 0.311 & 0.272 & 0.311 & 0.272 & 0.321  \\
& 720 & \textbf{0.331} & \textbf{0.339} & 0.343 & 0.351 & 0.349 & 0.351 & 0.353 & 0.348 & 0.362 & 0.366   \\
\cmidrule(lr){2-12}
 Degradation & & - & - & 4.41\% & 3.86\% & 5.53 \% & 4.55 \%  & 5.73\%  & 5.32 \% & 9.85 \% & 10.97 \%   \\
\midrule
\multirow{4}{*}{Etth2} 
& 96  & \textbf{0.282} & \textbf{0.336} & 0.292 & 0.352 & 0.295 & 0.356 & 0.296 & 0.358 & 0.312 & 0.377   \\
& 192 & \textbf{0.361} & \textbf{0.387} & 0.376 & 0.399 & 0.372 & 0.401 & 0.376 & 0.403 & 0.395 & 0.412   \\
& 336 & \textbf{0.395} & \textbf{0.421} & 0.411 & 0.440 & 0.413 & 0.430 & 0.409 & 0.436 & 0.428 & 0.462   \\
& 720 & \textbf{0.411} & \textbf{0.426} & 0.429 & 0.441 & 0.426 & 0.437 & 0.431 & 0.439 & 0.452 & 0.468  \\
\cmidrule(lr){2-12}
Degradation&  & - & - & {4.04 \%} & 3.97 \% & 3.97 \% & 3.57\% & 4.38\% & 4.32\% & 9.60 \% & 9.56\% \\
\bottomrule
\end{tabular}
}

\end{table*}

\begin{enumerate}

    \item {Efficacy of Cross-Scale Mixing (`w/o Mixer`):} Instead of Cross Scale Mixing:
    Removing cross scale mixing mechanism and replace it with simple feature average will lead to largest performance drop. The Weather's MSE increases by 9.85\% and ETTh2's MSE increases by 9.60\%. It shows that learning certain structured interaction among different temporal scales is vital to construct coherent multi-scale representation

    \item {Efficacy of Gated Fusion (`w/o Gating`):}  Using direct additive fusion in place of gated fusion degrades the model performance (MSE increases by 5.73\% on Weather and 4.38\% on ETTh2). It is validated that gate is able to control the information flow and enhance relevant frequency component while suppressing unnecessary noise.
    
    \item {Efficacy of Wavelet Representation (`w/o Wavelet`):} `w/o Wavelet` removes the whole wavelet stream. The model performance degrades by 5.53\% on Weather and 3.97\% on ETTh2. The above results further prove the effectiveness of wavelet-based time-frequency analysis. Especially for non-stationary cases, wavelet method can provide precise temporal localization of frequency content.

    \item {Efficacy of Frequency Router (`w/o Router`):} If the adaptive frequency weighting router is removed, the model performance will be consistently suppressed (MSE increases by 4.41\% on Weather, 4.04\% on ETTh2). It is proved that adaptive selection of relevant frequency bands and dynamic suppression of irrelevant ones are also an important contribution to the final forecasting accuracy.
\end{enumerate}

As shown in above ablation experiments, the experiments have shown that all components are helpful and necessary to the final design. If remove one of them, the performance will be suppressed and decreased. The removal of one component can be regarded as the combination of other components. So these results also provide strong evidence for our design. Meanwhile, adaptive frequency process mechanism and multi-scale temporal modeling are also proved to be effective. 

\subsection{Sensitivity Analysis of Hyperparameters}

In this section, we perform sensitivity analysis on some important hyper-parameters of AWEMixer. This analysis, therefore, investigates how performance varies with changes to the core components of our architecture:  the number of fusion layers and the specifics of the wavelet decomposition. All sensitivity experiments are conducted on the ETTh2 and Weather datasets, with results averaged over all four prediction horizons ($T \in \{96, 192, 336, 720\}$).

\begin{figure}[h!]
    \centering

    \begin{minipage}[b]{0.50\columnwidth}
        \centering
        \includegraphics[width=\linewidth]{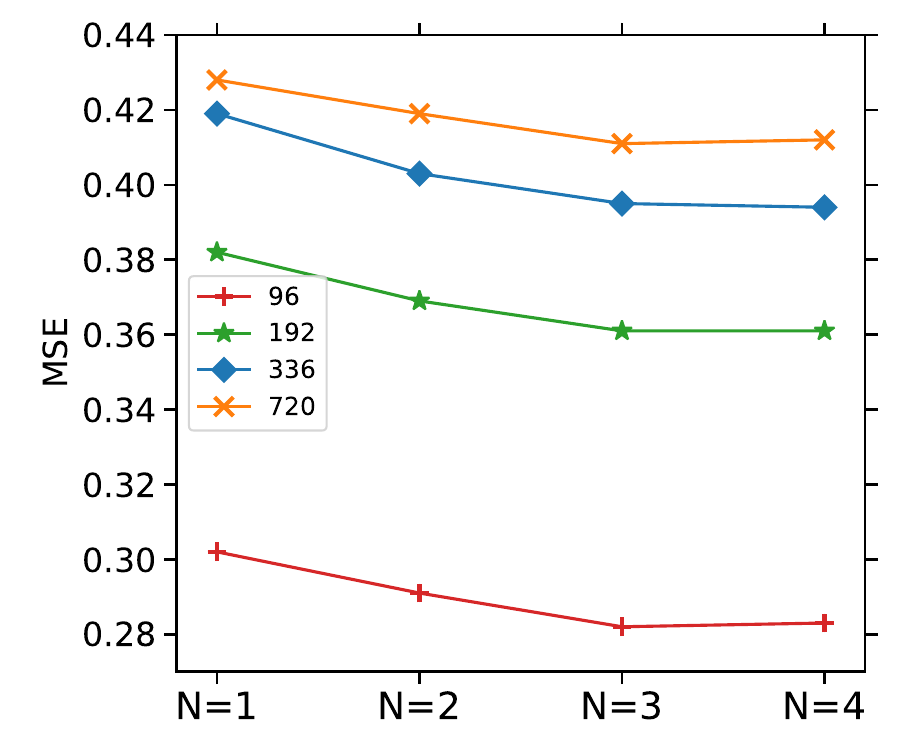}
        \centerline{(a) ETTh-2 Dataset}
    \end{minipage}
    \hfill   
    \begin{minipage}[b]{0.50\columnwidth}
        \centering
        \includegraphics[width=\linewidth]{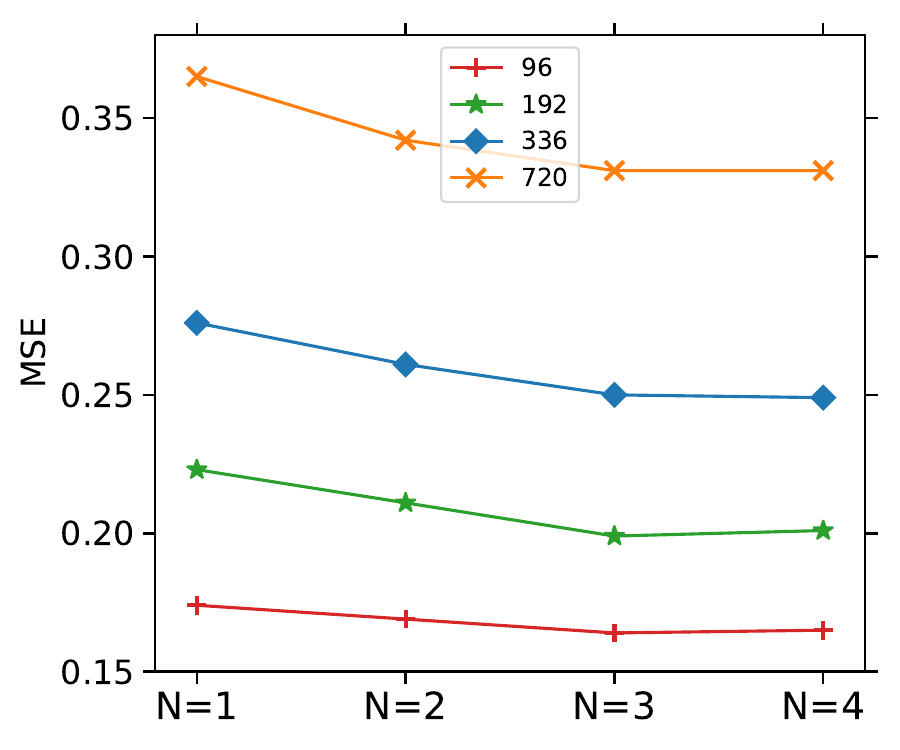}
        \centerline{(b) Weather Dataset}
    \end{minipage}

    \caption{Sensitivity analysis of the number of Gated Fusion layers ($N$). The plots show results on (a) the ETTh2 dataset and (b) the Weather dataset. Performance generally improves as $N$ increases from 1 to 3, with $N=3$ offering the best balance.}
    \label{fig:sensitivity_n_layers}
\end{figure}

The number of Gated Fusion Blocks, $N$, controls the depth of the time-frequency interaction. Each layer allows the temporal representation to be progressively refined with information from the wavelet pyramid. We experiment with a stack of $N \in \{1, 2, 3, 4\}$ fusion layers. The results are presented in Figure \ref{fig:sensitivity_n_layers}.
\subsubsection{Impact of Number of Fusion Layers }

The results show that performance improves significantly when moving from one layer to two, indicating that a single fusion step is insufficient to fully integrate the time-frequency information. The best performance is typically achieved with $N=2$ or $N=3$ layers. Beyond three layers, we observe a slight performance degradation, which can be attributed to the increased risk of overfitting and potential challenges in training a deeper network for this specific task. This finding validates our choice of $N=3$ for the main experiments and shows that a moderately deep fusion process is optimal.

\subsubsection{Impact of Wavelet Decomposition Level}

The wavelet decomposition level, $J$, determines the granularity of the frequency-domain analysis. A higher level provides a finer-grained decomposition, separating the signal into more frequency bands, but also results in a very short approximation coefficient array, which might lose crucial low-frequency information. We test the model's performance with the decomposition level varying as $J \in \{2, 3, 4, 5\}$.

\begin{figure}[h!]
    \centering

    \begin{minipage}[b]{0.48\columnwidth}
        \centering
        \includegraphics[width=\linewidth]{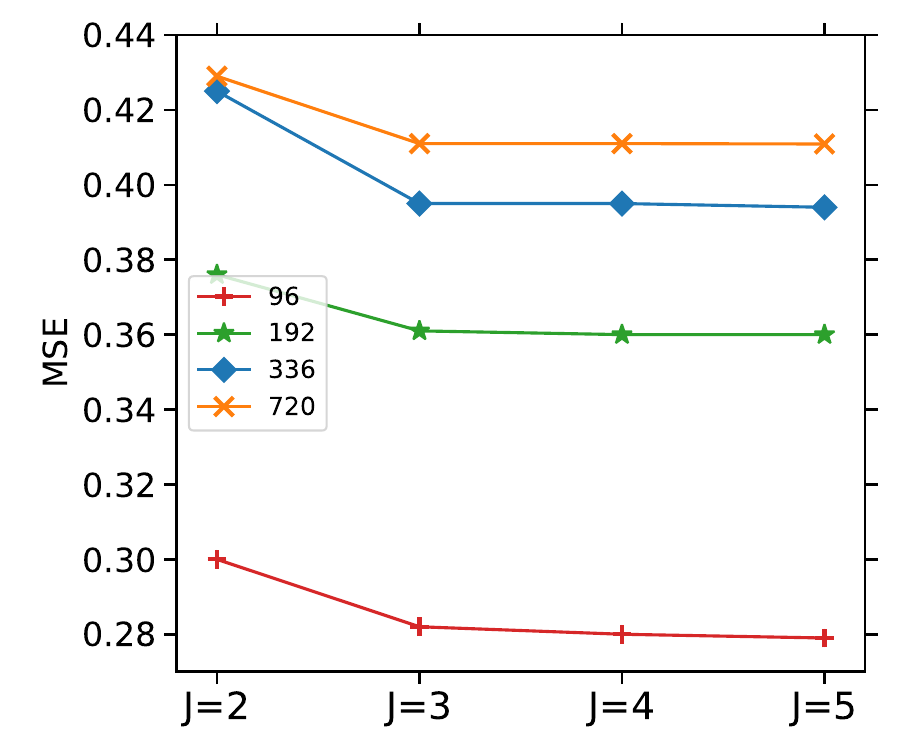}
        \centerline{(a) ETTh-2 Dataset}
    \end{minipage}
    \hfill   
    \begin{minipage}[b]{0.48\columnwidth}
        \centering
        \includegraphics[width=\linewidth]{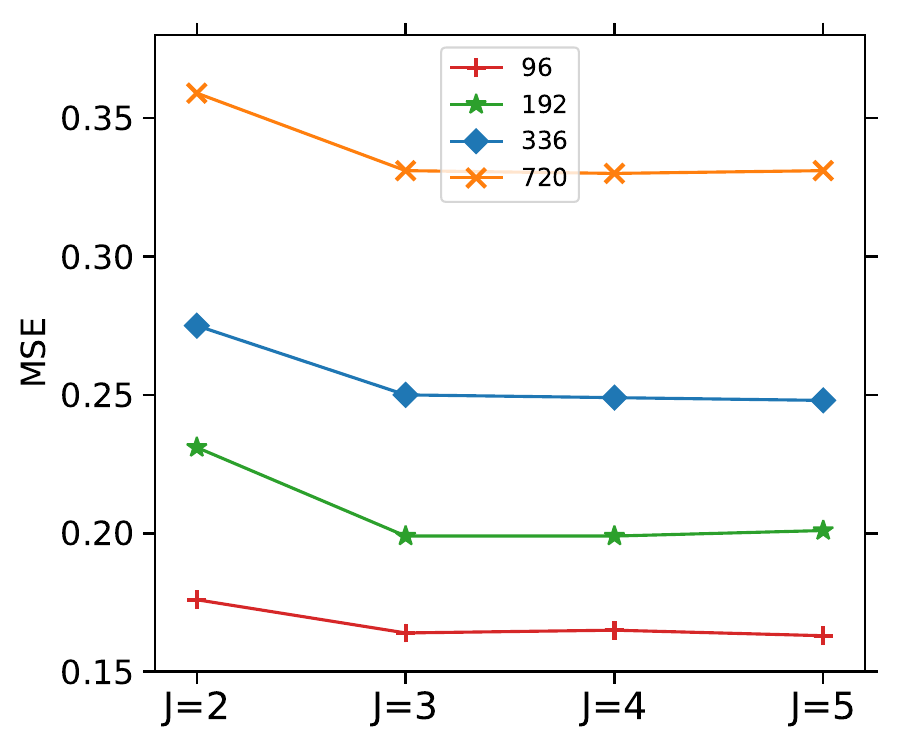}
        \centerline{(b) Weather Dataset}
    \end{minipage}

    \caption{Sensitivity analysis of the wavelet decomposition level $J$. Results are shown for (a) the ETTh2 dataset and (b) the Weather dataset. A level of 3 or 4 offers the best balance between decomposition granularity and preservation of trend information.}
    \label{fig:sensitivity_j_level}
\end{figure}

As shown in Figure \ref{fig:sensitivity_j_level}, the model performs best with a decomposition level of $J=3$ or $J=4$. When the level is too low ($J=2$), the frequency decomposition is too coarse, and the model struggles to separate complex seasonalities from the main trend. Conversely, when the level is too high ($J=5$), the performance also degrades. This is likely because a very high decomposition level results in an extremely short approximation coefficient ($cA_5$), which may be too compressed to accurately represent the long-term trend. This analysis confirms that a decomposition level of 3, as used in our primary experiments, is a robust and effective choice, providing a rich yet stable multi-band representation of the signal.

These sensitivity analyses show that although hyperparameter selection affects performance, AWEMixer is stable over reasonable variations in  fusion depth and wavelet decomposition level. This stability shows practical viability of the architecture in deployment scenarios where it may not be feasible to exhaustively tune these parameters.

\section{Discussion}
\label{sec:discussion}

The experimental results show that AWEMixer achieves consistent and strong performance across forecasting tasks. In this section, we provide a conceptual comparison to related state-of-the-art models, detail the key advancements brought by AWEMixer, and discuss the model's current limitations and future work directions.

\subsection{Comparison with Frequency-Domain Models}

The models FEDformer \cite{zhou2022fedformer} and TimesNet \cite{DBLP:conf/iclr/WuHLZ0L23} also possess a similar motivation to incorporate frequency-domain information. However, the approaches taken by these models are quite different. FEDformer performs attention on the Fourier domain, which is still a global operation and hence less sensitive to local variations. TimesNet converts one-dimensional sequences into two-dimensional tensors, allowing periodicity to be modelled via 2D convolutional kernels.

Different to a purely Fourier-based approach, AWEMixer uses the Wavelet Transform to acquire localized time–frequency representations. The model, represented by the Frequency Router, is then able to selectively and adaptively attend to information based on time-varying frequency characteristics. Unlike Fourier-based models, AWEMixer is able to capture both global periodicity and transient changes, allowing for more flexible and adaptive modelling of real-world periodic, multi-scale and non-stationary time series.

\subsection{Enhancement over the Mixer Paradigm}

As the name suggests, AWEMixer builds upon the efficient MLP-Mixer architecture. Previous Mixer-based models such as TimeMixer \cite{wang2024timemixer} and MSDS-Mixer \cite{MSDMixer} perform time-domain decomposition using methods such as downsampling and moving averages, and hence these models still need to implicitly learn frequency-related information in the temporal space.

In contrast, AWEMixer introduces an explicit, parallel frequency-processing stream coming from the Wavelet Transform. By allowing the localized frequency information to be dynamically fused with temporal features via a Gated Fusion mechanism, AWEMixer lightens the workload of the temporal mixer. Hence, the two-stream model is able to directly make use of structured frequency information instead of inferring it implicitly, facilitating better modelling of periodic, multi-scale and non-stationary patterns. This design choice is a key reason behind the model's strong empirical performance.

\subsection{Limitations and Future Work}

Despite the promising results, AWEMixer still has several limitations that can be improved via future work.
\begin{enumerate}

\item Fixed Mother Wavelet: Currently, AWEMixer uses a Haar mother wavelet. However, different time series may have different optimal mother wavelets. Therefore, future work could explore more diverse mother wavelets.

Current Limitation: Currently hard-coded to use a single mother wavelet (e.g., db4). Different datasets might have intrinsic periodic patterns that are better represented by different wavelet bases. Future work could investigate adaptive / learnable wavelet selection strategies.
\item {Global Routing Signal}:
Currently, Frequency Router computes the routing weights via a global FFT. When routing for highly non-stationary signals, i.e., the frequency components of a signal change quickly within a short time span, the router might benefit from being more adaptive to local frequency shifts. Alternative designs including windowed FFTs or routing metrics based on wavelet transforms could be explored.
\item {Finer-Grained Decomposition}: Our design is built upon DWT which has the advantage of being compact but coarse in decomposition. The Wavelet Packet Transform (WPT)  is finer in terms of frequency resolution and the output might capture more complicated oscillatory structures. Our approach could benefit from WPT when there is rich multi-frequency content in the signal.

\end{enumerate}
    
\section{Conclusion}
\label{con}
In this paper, we propose AWEMixer, an adaptive wavelet-enhanced mixing network for long-term forecasting of non-stationary time series. Particularly, we design a Frequency Router to adaptively weight different wavelet subbands in a globally periodic manner, and a Coherent Gated Fusion block to selectively inject more refined frequency components into temporal representations. Hence, our two-stream design can achieve fine-grained time-frequency localization through adaptive cross-domain interaction.

Empirical evaluations across seven public benchmarks demonstrate that AWEMixer achieves consistent improvements over state-of-the-art models, including recent Transformer- and MLP-based approaches. Ablation studies confirm the importance of both adaptive frequency weighting and gated fusion. Future work may explore learning wavelet representations end-to-end, incorporating exogenous variables, or extending the framework to related tasks such as anomaly detection and classification, where joint time-frequency analysis offers similar advantages.


%





\ifCLASSOPTIONcaptionsoff
  \newpage
\fi



%



\bibliographystyle{IEEEtran}
\bibliography{AWEMixer}
%

\begin{IEEEbiography}
[{\includegraphics[width=1in,height=1.25in,clip,keepaspectratio]{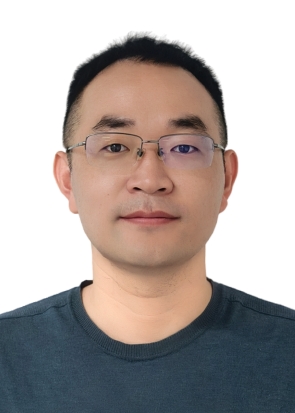}}]{Qianyang Li}
received the B.S. and M.S. degree from the Department of Mathematics, Harbin Institute of Technology, Weihai, China, in 2008 and 2010. He is currently a PhD candidate with the Computer science and technology School, Xi'an Jiaotong University. His main research interests include deep learning, data mining and time series.
\end{IEEEbiography}
\vspace{-1cm} 
\begin{IEEEbiography}
[{\includegraphics[width=1in,height=1.25in,clip,keepaspectratio]{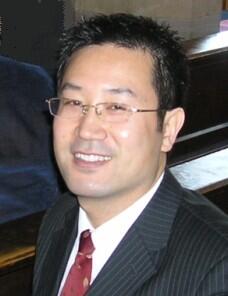}}]
{Xingjun Zhang} received his Ph.D degree in Computer Architecture from Xi'an Jiaotong University, China, in 2003. From Jan.2004 to Dec.2005, he was Postdoctoral Fellow at the Computer School of Beihang University, China. From Feb.2006 to Jan.2009, he was Research Fellow in the Department of Electronic Engineering of Aston University, United Kingdom. He is now a Full Professor and the Dean of the School of Computer Science \& Technology, Xi'an Jiaotong University. His research interests include high performance computing, big data storage system and machine learning acceleration.
\end{IEEEbiography}
\vspace{-1cm} 
\begin{IEEEbiography}[{\includegraphics[width=1in,height=1.25in,clip,keepaspectratio]{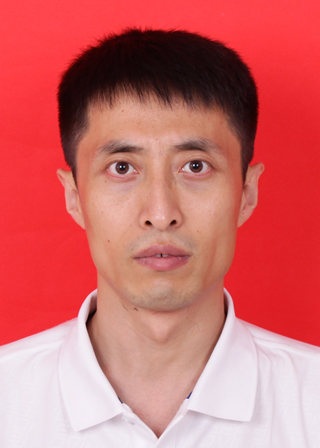}}]{Peng Tao}
received the B.S. degree in mechatronic engineering from Shandong Institute of Architecture and Engineering, China, in 2000. He received the M.S. degree in systems and application engineering and the Ph.D. degree in Systems Engineering from Beijing Institute of Technology, China, in 2005 and 2008, respectively. Since 2008, he has been with Shandong New Beiyang Information Technology Co., Ltd., where he is currently the Technical Director. His main research interests include pattern recognition, image detection, and data mining.
\end{IEEEbiography}
\vspace{-1cm} 
\begin{IEEEbiography}
[{\includegraphics[width=1in,height=1.25in,clip,keepaspectratio]{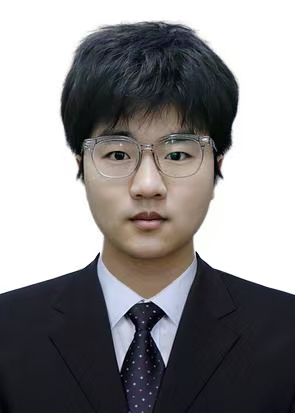}}]
{Shaoxun Wang} received the M.S. degree in Electronic Information from the School of Communication and Information Engineering, Xi'an University of Posts and Telecommunications, Xi'an, China, in 2024. He is currently pursuing the Ph.D. degree in Computer Science and Technology with the School of Computer Science and Technology, Xi'an Jiaotong University, Xi'an, China. 
\end{IEEEbiography}
\vspace{-1cm} 
\begin{IEEEbiography}
[{\includegraphics[width=1in,height=1.25in,clip,keepaspectratio]{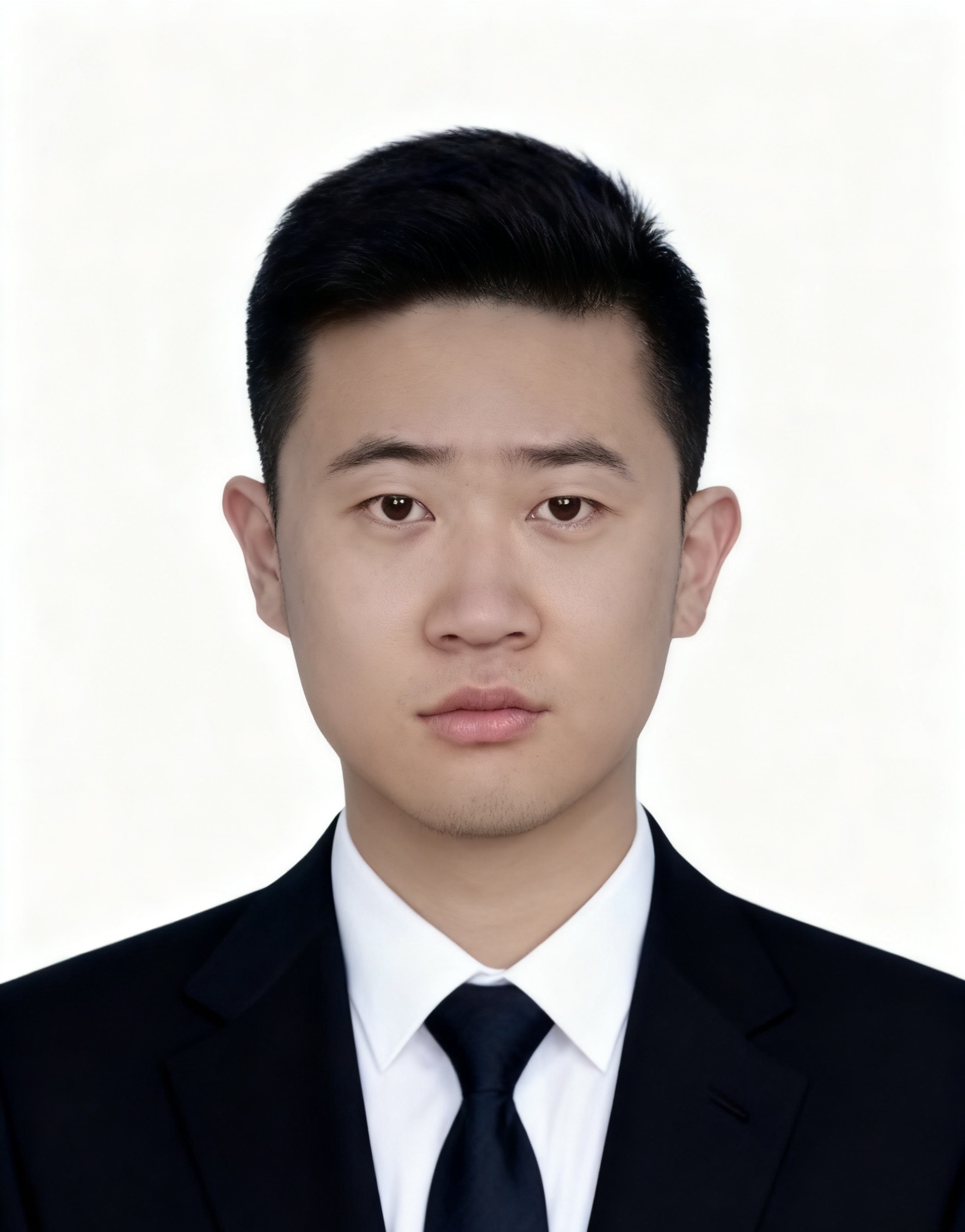}}]
{Yancheng Pan}  received the B.S. degree from the School of Cyber Engineering, Xidian University, in 2020. He is currently a PhD candidate with the School of Computer Science and Technology, Xi'an Jiaotong University, majoring in Computer Science and Technology.
\end{IEEEbiography}
\vspace{-1cm} 
\begin{IEEEbiography}
[{\includegraphics[width=1in,height=1.25in,clip,keepaspectratio]{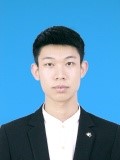}}]
{Jia Wei} received the Ph.D. degree from the School of Computer science and technology, Xi'an Jiaotong University, Xi'an, China, in 2024. He is currently a  Postdoctoral Fellow with the department of Computer science and technology , Tsinghua University. His research interests include artificial intelligent systems, computer architecture, and deep learning.
\end{IEEEbiography}








\end{document}